\documentclass{article}

\usepackage{microtype}
\usepackage{graphicx}
\usepackage{subfigure}
\usepackage{booktabs} 
\usepackage{amssymb}
\usepackage{amsmath}
\usepackage{bbm}
\usepackage{xr}
\usepackage{hyperref}
\usepackage{arydshln}

\usepackage[accepted]{icml2020}

\icmltitlerunning{The continuous categorical: a novel simplex-valued exponential family}

\begin{document}

\twocolumn[
\icmltitle{The continuous categorical: a novel simplex-valued exponential family}



\icmlsetsymbol{equal}{*}

\begin{icmlauthorlist}
\icmlauthor{Elliott Gordon-Rodriguez}{col}
\icmlauthor{Gabriel Loaiza-Ganem}{layer6}
\icmlauthor{John P. Cunningham}{col}
\end{icmlauthorlist}

\icmlaffiliation{col}{Department of Statistics, Columbia University}
\icmlaffiliation{layer6}{Layer 6 AI}

\icmlcorrespondingauthor{Elliott Gordon-Rodriguez}{eg2912@columbia.edu}
\icmlcorrespondingauthor{Gabriel Loaiza-Ganem}{gabriel@layer6.ai}
\icmlcorrespondingauthor{John P. Cunningham}{jpc2181@columbia.edu}

\icmlkeywords{Machine Learning, ICML}

\vskip 0.3in
]



\printAffiliationsAndNotice{}  

\begin{abstract}
Simplex-valued data appear throughout statistics and machine learning, for example in the context of transfer learning and compression of deep networks.
Existing models for this class of data rely on the Dirichlet distribution or other related loss functions; here we show these standard choices suffer systematically from a number of limitations, including bias and numerical issues that frustrate the use of flexible network models upstream of these distributions.
We resolve these limitations by introducing a novel exponential family of distributions for modeling simplex-valued data -- the \emph{continuous categorical}, which arises as a nontrivial multivariate generalization of the recently discovered continuous Bernoulli.
Unlike the Dirichlet and other typical choices, the continuous categorical results in a well-behaved probabilistic loss function that produces unbiased estimators, while preserving the mathematical simplicity of the Dirichlet.
As well as exploring its theoretical properties, we introduce sampling methods for this distribution that are amenable to the reparameterization trick, and evaluate their performance.
Lastly, we demonstrate that the continuous categorical outperforms standard choices empirically, across a simulation study, an applied example on multi-party elections, and a neural network compression task.\footnote{Our code is available at \url{https://github.com/cunningham-lab/cb_and_cc}}
\end{abstract}

\section{Introduction}

Simplex-valued data, commonly referred to as \emph{compositional data} in the statistics literature, are of great practical relevance across the natural and social sciences (see \citet{pawlowsky2006compositional, pawlowsky2011compositional, pawlowsky2015modeling} for an overview).
Prominent examples appear in highly cited work ranging from geology \cite{pawlowsky2004geostatistical, buccianti2006compositional}, chemistry \cite{buccianti2005new}, microbiology \cite{gloor2017microbiome}, genetics \cite{quinn2018understanding}, psychiatry \cite{gueorguieva2008dirichlet}, ecology \cite{douma2019analysing}, environmental science \cite{filzmoser2009univariate}, materials science \cite{na2014compositional}, political science \cite{katz1999statistical}, public policy \cite{breunig2012fiscal}, economics \cite{fry2000compositional}, and the list goes on. 
An application of particular interest in machine learning arises in the context of model compression, where the class probabilities outputted by a large model are used as `soft targets' to train a small neural network \cite{bucilua2006model,ba2014deep,hinton2015distilling}, an idea used also in transfer learning \cite{tzeng2015simultaneous, parisotto2015actor}.

The existing statistical models of compositional data come in three flavors.
Firstly, there are models based on a Dirichlet likelihood, for example the \emph{Dirichlet GLM} \cite{campbell1987multivariate, gueorguieva2008dirichlet, hijazi2009modelling} and \emph{Dirichlet Component Analysis} \cite{wang2008dirichlet, masoudimansour2017dirichlet}.
Secondly, there are also models based on applying classical statistical techniques to $\mathbb R^K$-valued transformations of simplex-valued data, notably via \emph{Logratios} \cite{aitchison1982statistical,  aitchison1994principles, aitchison1999logratios, egozcue2003isometric, ma2016decorrelation, avalos2018representation}.
Thirdly, the machine learning literature has proposed predictive models that forgo the use of a probabilistic objective altogether and optimize the categorical cross-entropy instead \cite{ba2014deep,hinton2015distilling,sadowski2018neural}.
The first type of model, fundamentally, suffers from an ill-behaved loss function (amongst other drawbacks) (\S\ref{sec:background}), which constrain the practitioner to using inflexible (linear) models with only a small number of predictors.
The second approach typically results in complicated likelihoods, which are hard to analyze and do not form an exponential family, forgoing many of the attractive theoretical properties of the Dirichlet. 
The third approach suffers from ignoring normalizing constants, sacrificing the properties of maximum likelihood estimation (see \citet{loaiza2019continuous} for a more detailed discussion of the pitfalls).

In this work, we resolve these limitations simultaneously by defining a novel exponential family supported on the simplex, the \emph{continuous categorical} (CC) distribution.
Our distribution arises naturally as a multivariate generalization of the recently discovered continuous Bernoulli (CB) distribution \cite{loaiza2019continuous}, a $[0,1]$-supported exponential family motivated by Variational Autoencoders \citep{kingma2013auto}, which showed empirical improvements over the beta distribution for modeling data which lies close to the extrema of the unit interval.

Similarly, the CC will provide three crucial advantages over the Dirichlet and other competitors: it defines a coherent, well-behaved and computationally tractable log-likelihood (\S\ref{sec:conv}, \ref{sec:normalizing}), which does not blow up in the presence of zero-valued observations (\S\ref{sec:extrema}), and which produces unbiased estimators (\S\ref{sec:exp_fam}).
Moreover, the CC model presents no added complexity relative to its competitors; it has one fewer parameter than the Dirichlet and a normalizing constant that can be written in closed form using elementary functions alone (\S\ref{sec:normalizing}). 
The continuous categorical brings probabilistic machine learning to the analysis of compositional data, opening up avenues for future applied and theoretical research.




\section{Background} \label{sec:background}

Careful consideration of the Dirichlet clarifies its shortcomings and the need for the CC family.  
The Dirichlet distribution is parameterized by $ \text{\boldmath$\alpha$} \in \mathbb R_+^K $ and defined on the simplex  $\mathbb{S}^{K-1} = \{ \bold x \in \mathbb R_+^{K-1} : \sum_{i=1}^{K-1} x_i < 1 \} $ by:
\begin{align} \label{eq:dirichlet}
p(x_1,\dots,x_{K-1} ; \text{\boldmath$\alpha$} ) = 
\frac{1}{B(\text{\boldmath$\alpha$}) } \prod_{i=1}^K x_i^{\alpha_i - 1},
\end{align}
where $B(\text{\boldmath$\alpha$})$ denotes the multivariate beta function, and $x_K = 1 - x_1 - \cdots - x_{K-1}$.\footnote{Note that the $K$th component does not form part of the argument; it is a deterministic function of the $(K-1)$-dimensional random variable $\bold x$.}

This density function presents an attractive combination of mathematical simplicity, computational tractability, and the flexibility to model both interior modes and sparsity, as well as defining an exponential family of distributions that provides a multivariate generalization of the beta distribution and a conjugate prior to the multinomial distribution.
As such, the Dirichlet is by far the best known and most widely used probability distribution for modeling simplex-valued random variables \cite{ng2011dirichlet}.
This includes both the latent variables of mixed membership models \cite{erosheva2002grade, blei2003latent, barnard2003matching}, and the statistical modeling of compositional outcomes \cite{aitchison1982statistical, campbell1987multivariate, hijazi2003analysis, wang2008dirichlet}.
Our work focuses on the latter, with special attention to the setting where we aim to learn a (possibly nonlinear) regression function that models a simplex-valued response in terms of a (possibly large) set of predictors.
In this context, and in spite of its strengths, the Dirichlet distribution suffers from three fundamental limitations.

First, flexibility: 
while the ability to capture interior modes might seem intuitively appealing, in practice it thwarts the optimization of predictive models of compositional data (as we will demonstrate empirically in section \S\ref{sec:election}).
To illustrate why, consider fitting a Dirichlet distribution to a single observation lying inside the simplex. Maximizing the likelihood will lead to a point mass on the observed datapoint (as was also noted by \citet{sadowski2018neural}); the log-likelihood will diverge, and so will the parameter estimate (tending to $\infty$ along the line that preserves the observed proportions).
This example may seem degenerate; after all, any dataset that would warrant a probabilistic model had better contain more than a single observation, at which point no individual point can `pull' the density onto itself.
However, in the context of predictive modeling, observations \emph{will} often present unique input-output pairs, particularly if we have continuous predictors, or a large number of categorical ones.
Thus, any regression function that is sufficiently flexible, such as a deep network, or that takes a sufficiently large number of inputs, can result in a divergent log-likelihood, frustrating an optimizer's effort to find a sensible estimator.
This limitation has constrained Dirichlet-based predictive models to the space of linear functions \cite{campbell1987multivariate, hijazi2009modelling}.

Second, tails: the Dirichlet density diverges or vanishes at the extrema for all but a set of measure zero values of $\text{\boldmath$  \alpha$}$, so that the log-likelihood is undefined whenever an observation contains zeros (transformation-based alternatives, such as Logratios, suffer the same drawback).
However, zeros are ubiquitous in real-world compositional data, a fact that has lead to the development of numerous hacks, such as \citet{palarea2008modified, scealy2011regression, stewart2011managing, hijazi2011algorithm, tsagris2018dirichlet}, each of which carries its own tradeoffs.

Third, bias: an elementary result is that the MLE of an exponential family yields an unbiased estimator for its sufficient statistic.
However, the sufficient statistic of the Dirichlet distribution is the logarithm of the data, so that by Jensen's inequality, the MLE for the mean parameter, which is often the object of interest, is biased.
While the MLE is also asymptotically consistent, this is only the case in the unrealistic setting of a true Dirichlet data-generating process, as we will illustrate empirically in section \S\ref{sec:dirichlet_bias}.

\section{The continuous categorical distribution}\label{sec:cc}

These three limitations motivate the introduction of a novel exponential family, the \emph{continuous categorical} (CC), which is defined on the closed simplex, $\text{cl}( \mathbb{S}^{K-1} ) =  \{ \bold x \in \mathbb R^{K-1} : \bold x \ge \boldsymbol 0, \sum_{i=1}^{K-1} x_i \le 1 \} $, by:
\begin{align} \label{eq:cc_def}
\bold x \sim \mathcal{CC}(\text{\boldmath$\lambda$}) \iff p(x_1,\dots,x_{K-1} ; \text{\boldmath$\lambda$}) \propto \prod_{i=1}^K \lambda_i^{x_i},
\end{align}
where, again, $x_K = 1 - x_1 - \cdots - x_{K-1}$.
Without loss of generality we restrict the parameter values $\{ \text{\boldmath$\lambda$} \in \mathbb R_+^K : \sum_i \lambda_i = 1\}$; such a choice makes the model identifiable.
The CC density looks much like the Dirichlet, except that we have switched the role of the parameter and the variable.
However, this simple exchange results in a well-behaved log-likelihood that can no longer concentrate mass on single interior points (\S \ref{sec:conv}), nor diverge at the extrema (\S\ref{sec:extrema}), and that produces unbiased estimators (\S\ref{sec:exp_fam}).

\subsection{Convexity and modes} \label{sec:conv}

\begin{figure*} 
\centering
  \includegraphics[width=.2\linewidth]{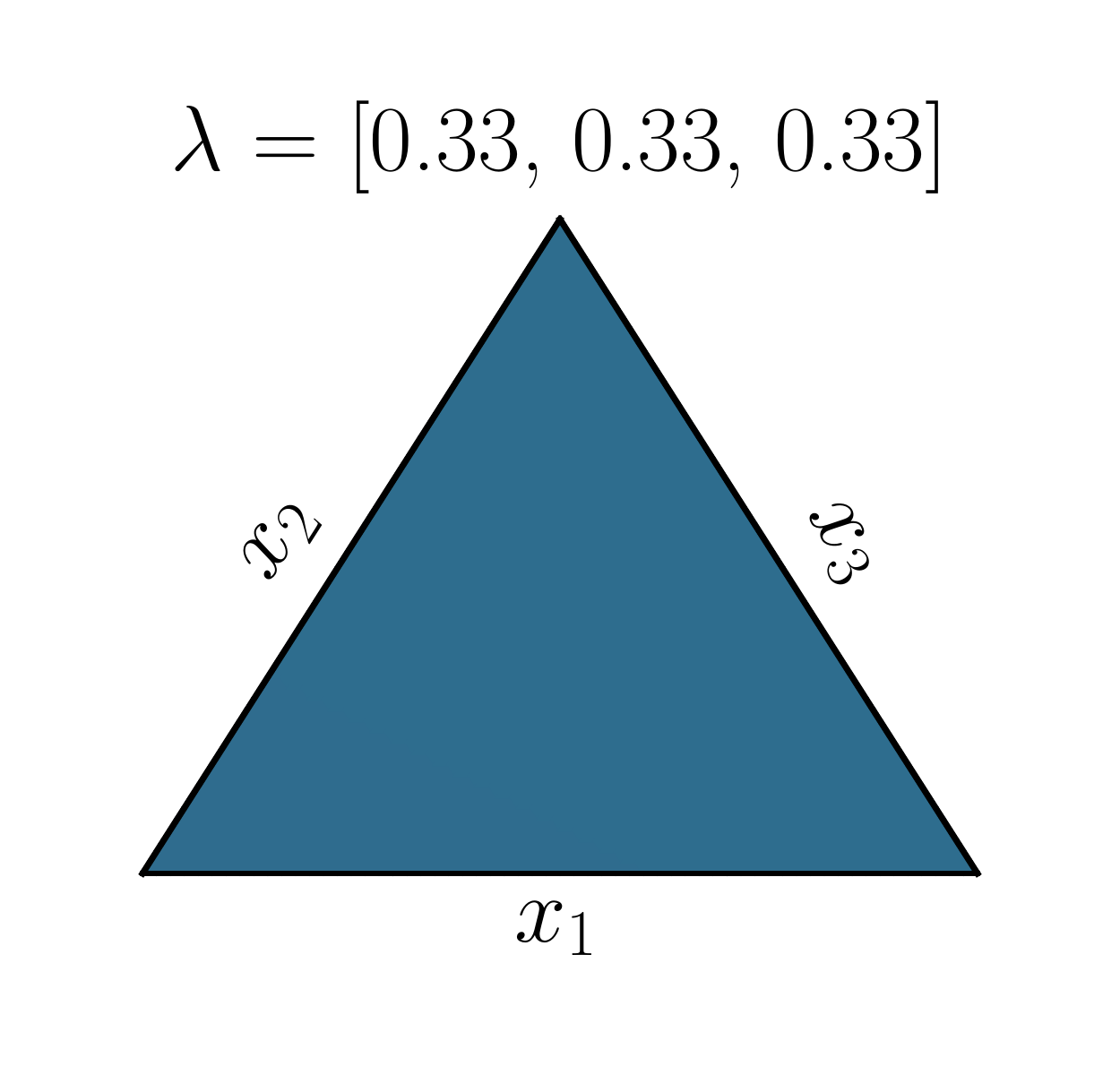}
  \includegraphics[width=.2\linewidth]{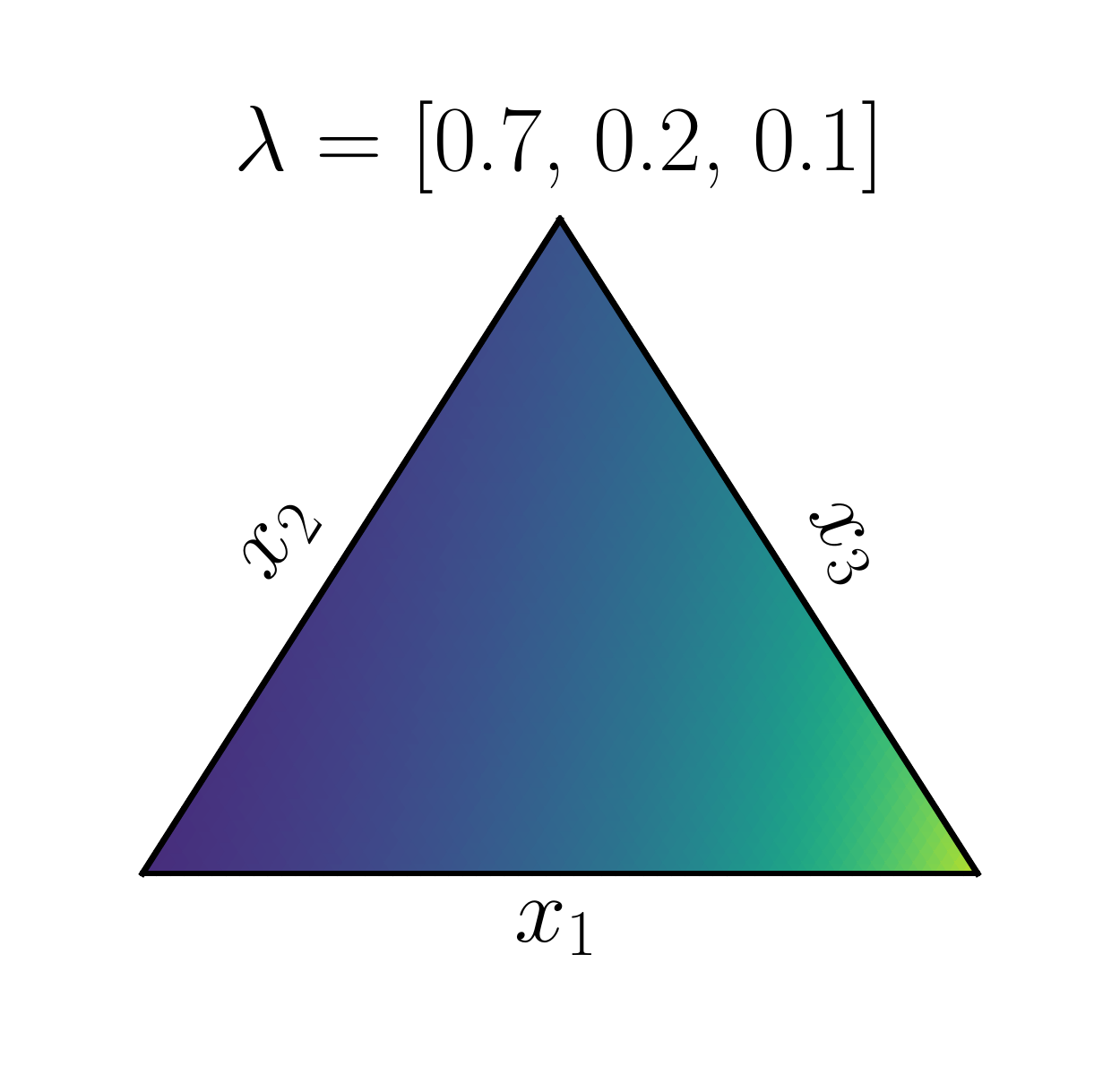}
  \includegraphics[width=.21\linewidth]{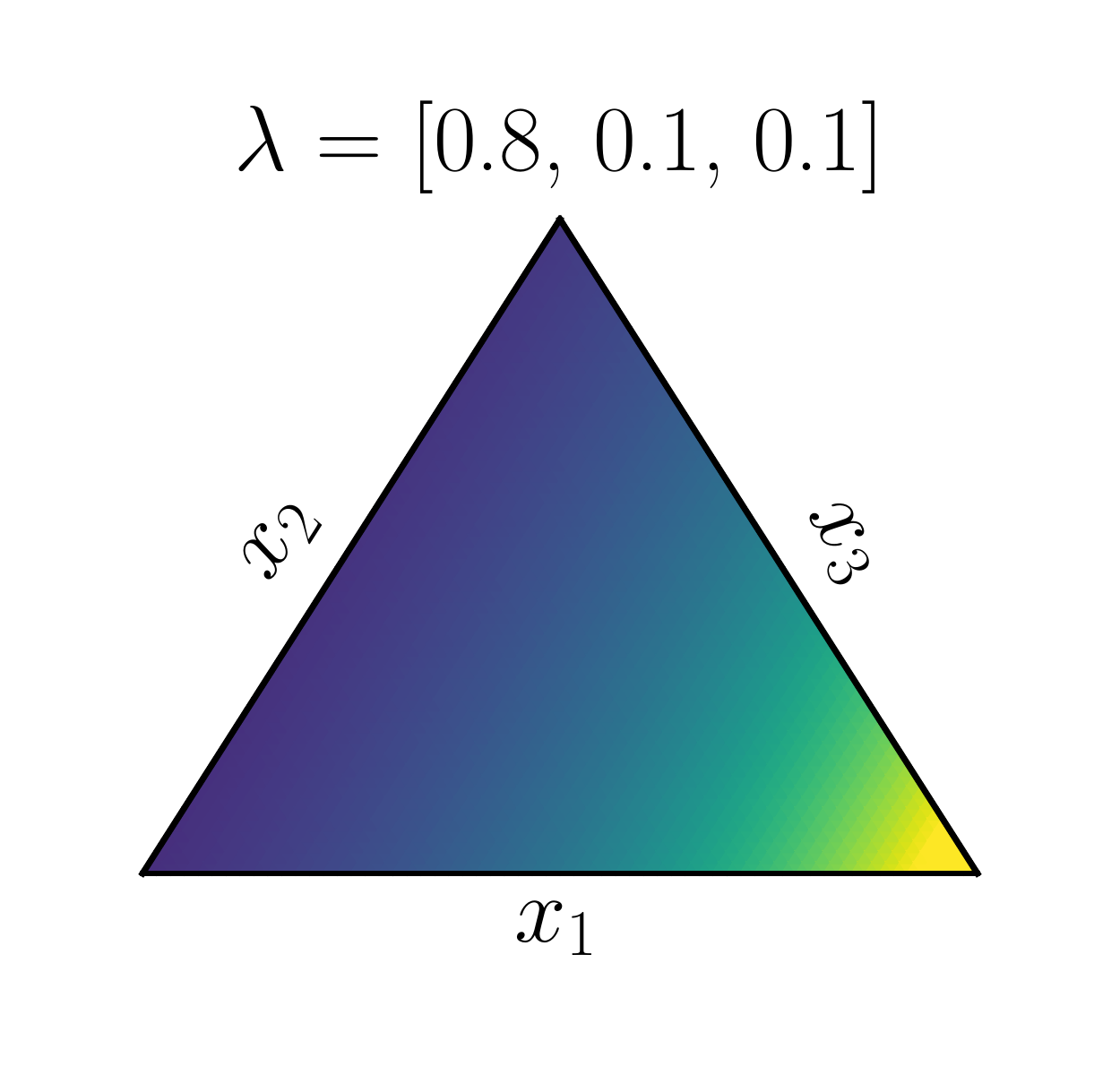}
  \includegraphics[width=.21\linewidth]{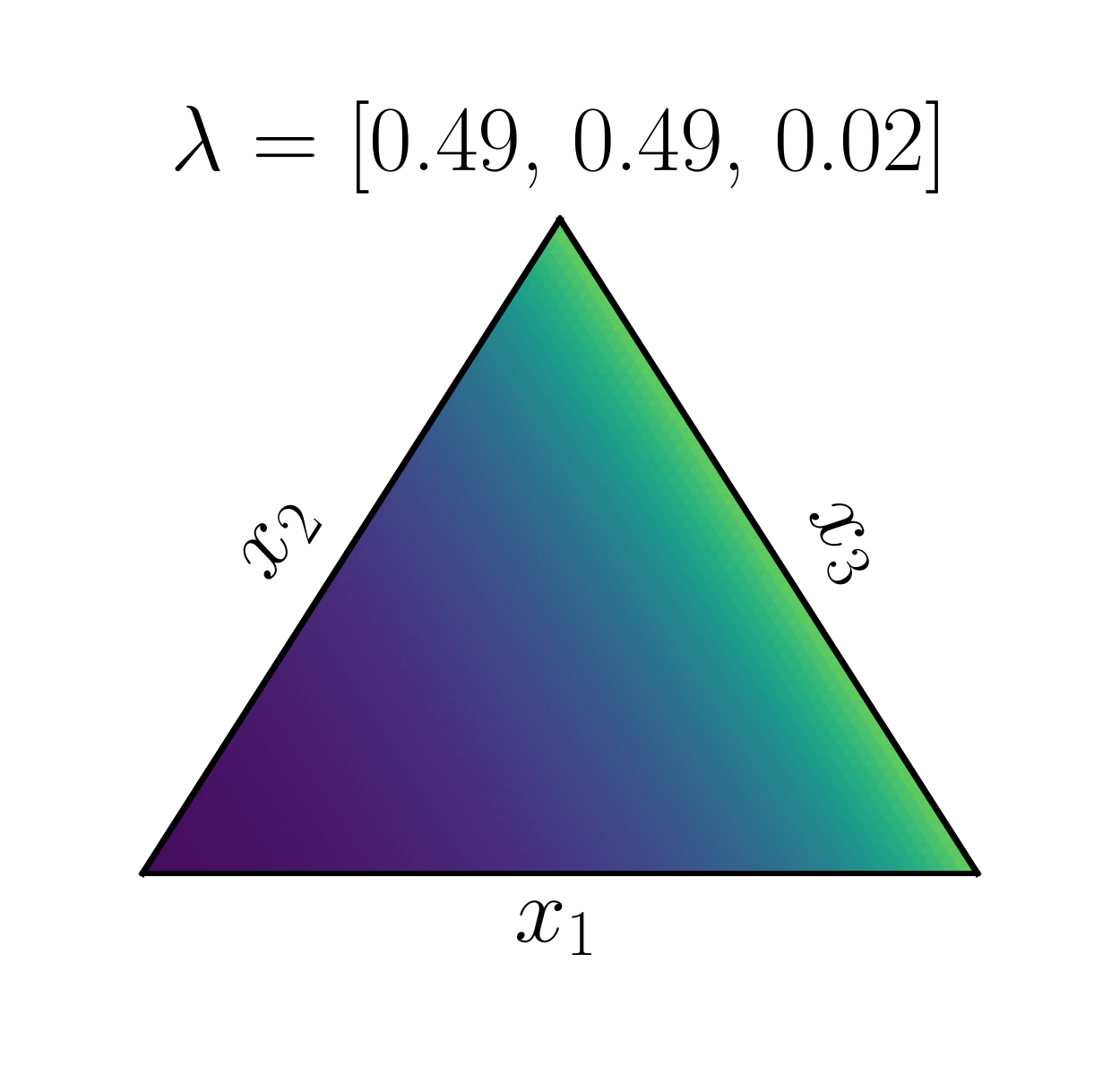}
  \includegraphics[width=.08\linewidth]{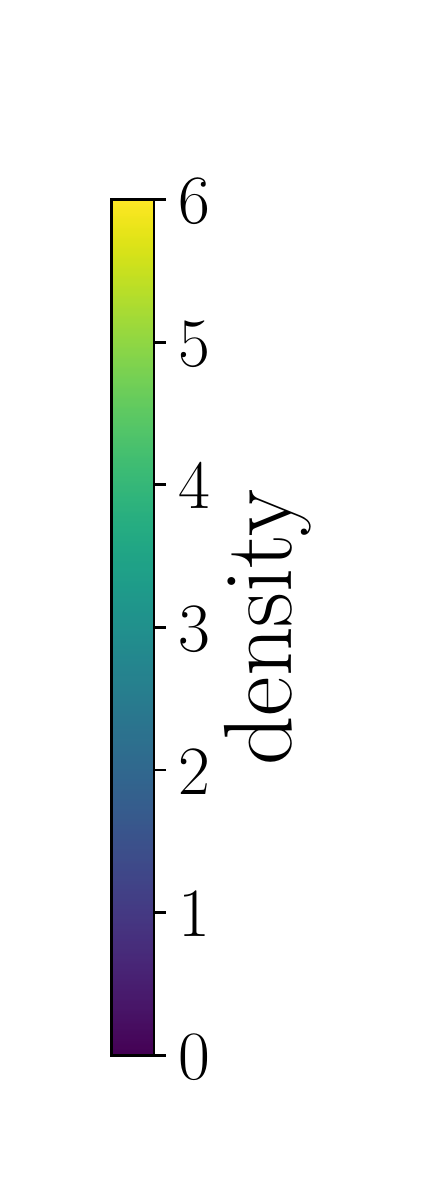}
  \caption{Density heatmaps of the 2-dimensional CC (defined on the 3-simplex). We show a near-uniform example, followed by more extremal examples, as well as a bimodal example (with the modes necessarily at the extrema). Note that, while we have defined the CC distribution on the space $\{ \bold x \ge \boldsymbol 0 : \sum_{i=1}^{K-1} x_i \le 1\}$, we plot the density on the equivalent set $\{ \bold x \ge \boldsymbol 0 : \sum_{i=1}^K x_i = 1\}$. } \label{fig:densities}
\end{figure*}

Because the CC log-likelihood is linear in the data (equation \ref{eq:cc_def}), the density is necessarily convex.
It follows that the modes of the CC are at the extrema; example density plots are shown in figure \ref{fig:densities}.
In this sense, the CC is a less flexible family than the Dirichlet, as it cannot represent interior modes.
In the context of fitting a probability distribution (possibly in a regression setting) to compositional outcomes, however, this choice prevents the CC from concentrating mass on single observations, a fundamental tradeoff with the Dirichlet distribution and other transformation-based competitors such as \citet{aitchison1982statistical,  aitchison1994principles, aitchison1999logratios}.
The mode of the CC is the basis vector associated with the $\text{argmax}(\lambda_1, \dots, \lambda_K)$ index of the data, provided the maximizer is unique.

\subsection{Concentration of mass} \label{sec:extrema}

The CC exhibits very different concentration of mass at the extrema relative to the Dirichlet.
The former is always finite and strictly positive, whereas the latter either diverges or vanishes for almost all parameter values, in other words:
\begin{align}
\lim_{x_j \to 0} \log \frac{CC(\bold x | \text{\boldmath$  \lambda$})}{Dirichlet(\bold x | \text{\boldmath$  \alpha$})} 
& \to
\begin{cases}
\infty, & \text{if } \alpha_j > 1
\\
-\infty, & \text{if } \alpha_j < 1
\end{cases}
.
\end{align}

The important distinction lies at the limit points; the CC is supported on the \emph{closed} simplex, whereas the Dirichlet and its transformation-based alternatives are defined only on its interior.
Thus, the CC log-likelihood can automatically model data with zeros, without requiring specialized techniques such as \citet{palarea2008modified, scealy2011regression, stewart2011managing, tsagris2018dirichlet, hijazi2011algorithm}, to name but a few.
The sheer amount of published work dedicated to this long-standing issue should convince the reader that this property of the CC distribution provides a substantial advantage in an applied modeling context.

\subsection{Exponential Family} \label{sec:exp_fam}

The CC defines an exponential family of distributions; noting that by definition $x_K = 1- x_1 - \cdots - x_{K-1}$, we can rewrite equation \ref{eq:cc_def}:

\begin{align}
p(\bold x; \text{\boldmath$\lambda$} ) \propto \exp\left( \sum_{i=1}^{K-1} x_i \log{\frac{\lambda_i}{\lambda_K}} \right) 
.
\end{align}

Letting $\eta_i = \log \frac{\lambda_i}{\lambda_K}$, so that $\lambda_i = \frac{\exp(\eta_i)}{\sum_{k=1}^K \exp(\eta_k) }$, gives the natural parameter of our exponential family.
Under this parameterization, we can ignore the $K$th component, $\eta_K = \log \frac{\lambda_K}{\lambda_K} \equiv 0$, and our parameter space becomes the unrestricted $\text{\boldmath$\eta$} \in \mathbb R ^{K-1}$, which is more convenient for optimization purposes and will be used throughout our implementation.
With a slight abuse of notation,\footnote{We will write $\bold x \sim \mathcal{CC}(\text{\boldmath$\lambda$})$ and $\bold x \sim \mathcal{CC}(\text{\boldmath$\eta$})$ interchangeably depending on context, and similarly for the density functions $p(\bold x; \text{\boldmath$\lambda$} ) $ and $p(\bold x; \text{\boldmath$\eta$} ) $.} our density simplifies to:
\begin{align} \label{eq:cc_nat}
p( x_1, \dots, x_{K-1}; \text{\boldmath$\eta$} ) \propto \exp(\text{\boldmath$\eta$}^\top \bold x).
\end{align}

This last formulation makes it apparent that, under a CC likelihood, the mean of the data is minimal sufficient.
By the standard theory of exponential families, this implies that the MLE of the CC distribution will produce an unbiased estimator of the mean parameter.
To be precise, if $\hat{\text{\boldmath$\lambda$}}$ maximizes the CC likelihood, and $\hat{\text{\boldmath$\mu$}}$ is the corresponding mean parameter obtained from $\hat{\text{\boldmath$\lambda$}}$, then $\hat{\text{\boldmath$\mu$}} = \bar{\bold x}$, where $\bar{\bold x}$ is the empirical average of the data.
Thus, the CC MLE is unbiased for the true mean, irrespective of the data-generating distribution.

This fact stands in contrast to the Dirichlet, in which the sufficient statistic is the mean of the logarithms, or other competitors, in which less can be said about the bias, partly as a result of their added complexities.
Not only are these biases undesirable at a philosophical level, but we will also find in section \ref{sec:experiments} that, empirically, they compromise the performance of the Dirichlet relative to the CC.

\subsection{Normalizing constant} \label{sec:normalizing}

For our new distribution to be of practical use, we must first derive its normalizing constant, which we denote by $C(\text{\boldmath$\eta$})$, defined by the equation:
\begin{align}
\int_{\mathbb{S}^{K-1}} C(\text{\boldmath$\eta$})
 \exp(\text{\boldmath$\eta$}^\top \bold x) d\mu(\bold x) = 1,
\end{align}
where $\mu$ is the Lebesgue measure.

\textbf{Proposition.} The normalizing constant of a $\mathcal{CC}(\text{\boldmath$\eta$})$ random variable is given by:
\begin{align}\label{eq:norm_const}
 C(\text{\boldmath$\eta$}) =
\left(
(-1)^{K+1} \sum_{k=1}^K 
\frac
 { { \exp\left(\eta_k\right)} }
 { \prod_{i \ne k } \left(\eta_i - \eta_k\right)}
\right)^{-1}
,
\end{align}
provided $\eta_i \ne \eta_k$ for all $i \ne k$.
\newline
\textbf{Proof.} We present a proof by induction in section A of the supplementary material.

Note it is only on a set of Lebesgue measure zero that $\eta_i = \eta_k$ for some $i \ne k$, and while the CC is still properly defined in these cases, the normalizing constant takes a different form.
However, evaluating equation \ref{eq:norm_const} when two (or more) parameters are approximately equal can result in numerical instability.
In our experiments (\S\ref{sec:experiments}), the limited instances of this issue were dealt with by zeroing out any error-inducing gradients during optimization.



It may come as a surprise that the normalizing constant of the CC distribution (\ref{eq:norm_const}) admits a simple closed form expression in terms of elementary functions; the same cannot be said of almost any multivariable function that one might hope to integrate over the simplex (compare to the Dirichlet density, which integrates to a product of gamma functions).
Thus, the CC turns out to be a very natural choice for a distribution on this sample space, which has not been previously proposed in the literature.
The appeal of equation \ref{eq:norm_const} is not just theoretical; it allows the CC log-likelihood to be optimized straightforwardly using automatic differentiation.
In other words, not only is the CC distribution able to address several key limitations of the Dirichlet, but it does so without sacrificing mathematical simplicity (the form of the densities are very similar) or exponential family properties, or adding additional model parameters.

\subsection{Mean and variance} \label{moments}

By the standard theory of exponential families, the mean and covariance of the CC distribution can be computed by differentiating the normalizing constant. 
Thus, while cumbersome to write down analytically, we can evaluate these quantities using automatic differentiation. 
Higher moments, including skewness and kurtosis, can also be derived from the normalizing constant, as well as a number of other distributional results, such as the characteristic function or KL divergence (see section B from the supplementary material).

\subsection{Related distributions} \label{sec:rel_dist}

The Dirichlet distribution can be thought of as a generalization of the beta distribution to higher dimensions, which arises by taking the product of independent beta densities and restricting the resulting function to the simplex.
In the same way, the CC provides a multivariate extension of the recently proposed continuous Bernoulli (CB) distribution \citep{loaiza2019continuous}, which is defined on  $[0,1]$ by:
\begin{align} 
x \sim \mathcal{CB}(\lambda) \iff p(x|\lambda) \propto \lambda^x (1-\lambda)^{1-x}
.
\end{align}
First, observe that this density is equivalent to the univariate case of the CC (i.e. $K=2$).
Second, note that the full CC density (\ref{eq:cc_def}) corresponds to the product of independent CB densities, restricted to the simplex, an idea that we will capitalize on when designing sampling algorithms for the CC (\S\ref{sec:samplers}).
In this sense, the beta, Dirichlet, CB and CC distributions form an intimately connected tetrad; the CB and the CC switch the role of the parameter and the variable in the beta and Dirichlet densities, respectively, and the Dirichlet and CC extend to the simplex the product of beta and CB densities, respectively.
The analogy to the beta and CB families is not just theoretical; the CB arose in the context of Variational Autoencoders \cite{kingma2013auto} and showed empirical improvements over the beta for modeling data which lies close to the extrema of the unit interval  \cite{loaiza2019continuous}.
Similarly, the CC provides theoretical and empirical improvements over the Dirichlet for modeling extremal data (\S\ref{sec:extrema}, \ref{sec:experiments}).

\section{Sampling}\label{sec:samplers}

Sampling is of fundamental importance for any probability distribution. We developed two novel sampling schemes for the CC, which we highlight here; full derivations and a study of their performance (including reparameterization gradients) can be found in the supplement (section C).
We note that, while the Dirichlet distribution can be sampled efficiently via normalized gamma draws or stick-breaking \cite{connor1969concepts}, we are not aware of any such equivalents for the CC (see section B.4 of the supplement).

Given the relationship between the CB and the CC densities (\S\ref{sec:rel_dist}), a naive rejection sampler for the CC follows directly by combining independent CB draws (using the closed form inverse cdf), and accepting only those that fall on the simplex.  
This basic sampler scales poorly in $K$.  
We improve upon it via a reordering operation, resulting in vastly better performance (see section C.2 and figure 1 from the supplement).  
The central concept of this scheme is to sort the parameter $\text{\boldmath$  \lambda$}$ and reject as soon as samples leave the simplex; this sampler is outlined in algorithm \ref{alg:ordered_rej}.

\begin{algorithm}
   \caption{Ordered rejection sampler}
   \label{alg:ordered_rej}
   {\bfseries Input:} target distribution $\mathcal{CC}(\text{\boldmath$  \lambda$})$.
   \\
   {\bfseries Output:} sample $\bold x$ drawn from target.
   \\ \vspace{-4mm}
   \begin{algorithmic}[1]
   \STATE Find the sorting operator $\pi$ that orders $\text{\boldmath$  \lambda$}$ from largest to smallest, and let  $\text{\boldmath$\tilde  \lambda$} = \pi( \text{\boldmath$  \lambda$})$.
   \STATE Set the cumulative sum $c \leftarrow 0$ and $i \leftarrow 2$.
   \WHILE{$c<1$ and $i\leq K$}
   \STATE Sample $x_{i} \sim \mathcal{CB}\left(\tilde{\lambda}_i /(\tilde{\lambda}_i + \tilde{\lambda}_1) \right)$.
   \STATE Set $c \leftarrow c + x_{i}$ and 
   \STATE Set $i \leftarrow i+1$.
   \ENDWHILE
   \STATE If $c>1$, go back to step 2.   
   \STATE Set $x_1 = 1 - \sum_{i=2}^K x_i$.
   \STATE Return $\bold x = \pi^{-1}(x_1,\dots,x_K)$.
\end{algorithmic}
\end{algorithm}

Algorithm \ref{alg:ordered_rej} performs particularly well in the case that $\text{\boldmath$  \lambda$}$ is unbalanced, with a small number of components concentrating most of the mass.
However, its efficiency degrades under balanced configurations of $\text{\boldmath$  \lambda$}$; this shortcoming motivates the need for our second sampler (algorithm \ref{alg:perm_sampling}), which performs particularly well in the balanced setting.  Conceptually, this second sampler will exploit a permutation-induced partition of the unit cube into simplices, combined with a relaxation of the CC which is invariant under permutations, and can be mapped back to the original CC distribution.

At a technical level, first note that if $\sigma$ is a permutation of $\{1,\dots,K-1\}$ and $\mathcal{S}_\sigma = \{\bold x \in \mathbb{R}^{K-1}: 0\leq x_{\sigma(1)} \leq x_{\sigma(2)} \leq \dots \leq x_{\sigma(K-1)} \leq 1\}$, then we can (almost, in the measure-theoretic sense) partition $[0,1]^{K-1} = \bigcup\limits \mathcal{S}_{\sigma}$ where the union is over all permutations.
Second, we can generalize the CC to an arbitrary sample space by writing $p_{\Omega} (\bold x| \boldsymbol \eta) \propto \exp(\boldsymbol \eta^\top \bold x) \mathbbm{1} (\bold x \in \Omega)$.
By the change of variable formula, we note that  this family is invariant to invertible linear maps in the sense that, if $\bold x \sim p_{\mathcal{A}}(\bold x | \boldsymbol \eta)$ and $\bold y =Q \bold x$, where $Q$ is an invertible matrix, then $\bold y \sim p_{Q(\mathcal{A})}(\bold y | \boldsymbol{ \tilde \eta})$, where $ \boldsymbol{ \tilde \eta} = Q^{-\top} \boldsymbol \eta$.
Hence, if $B$ is a lower-triangular matrix of ones and $id$ denotes the identity permutation, then $\mathcal{S}_{id} = B(\text{cl}(\mathbb{S}^{K-1}))$, and it follows that we can sample from $\bold x \sim \mathcal{CC}( \boldsymbol \eta)$ by sampling $\bold y \sim p_{\mathcal{S}_{id}}(\bold y | \boldsymbol{ \tilde \eta})$ and transforming with $\bold x = B^{-1} \bold y$.
Conveniently, $\bold y \sim p_{\mathcal{S}_{id}}(\bold y | \boldsymbol{ \tilde \eta})$ can be sampled by first drawing a $[0,1]^{K-1}$-valued vector of independent CB variates, namely $\bold y' \sim p_{[0,1]^{K-1}}(\bold y' | \boldsymbol{ \tilde \eta})$, then transforming into $\mathcal{S}_{id}$ by applying $\bold y = P \bold y'$, where $P$ is the permutation matrix that orders the elements of $\bold y'$, and finally accepting $\bold y$ with probability:
\begin{equation}\label{eq:acc_prob}
\alpha(\bold y, \boldsymbol{ \tilde \eta}, P) = \dfrac{p_{\mathcal{S}_{id}}(\bold y| \boldsymbol{\tilde \eta})}{\kappa(\boldsymbol{\tilde \eta},P)p_{\mathcal{S}_{id}}(\bold y| P^{-\top} \boldsymbol{\tilde \eta})}
,
\end{equation}
where $\kappa(\boldsymbol{\tilde \eta},P)$ is the rejection sampling constant:
\begin{align}
\kappa(\boldsymbol{\tilde \eta},P) = \max_{\bold y \in \mathcal{S}_{id}} \dfrac{p_{\mathcal{S}_{id}}(\bold y| \boldsymbol{\tilde \eta} )}{p_{\mathcal{S}_{id}}(\bold y| P^{-\top} \boldsymbol{\tilde \eta} )}
.
\end{align}

This sampling scheme is shown altogether in algorithm \ref{alg:perm_sampling}.
It performs particularly well in the case that  $\text{\boldmath$  \lambda$}$ is balanced, since the distributions induced on $\mathcal S_{id}$ by $\boldsymbol{\tilde \eta}$ and the permutations thereof, are similar, resulting in high acceptance probabilities.
Taken together with algorithm \ref{alg:ordered_rej}, our permutation sampler provides an efficient, theoretically understood, and reparameterizable sampling scheme for the CC.

\begin{algorithm} 
   \caption{Permutation sampler}
   \label{alg:perm_sampling}
   {\bfseries Input:} target distribution $\mathcal{CC}(\text{\boldmath$ \eta$})$.
   \\
   {\bfseries Output:} sample $\bold x$ drawn from target.
   \\ \vspace{-4mm}
   \begin{algorithmic}[1]
   \STATE Sample $\bold y' \sim p_{[0,1]^{K-1}}(\cdot | \boldsymbol{ \tilde \eta})$, where $\boldsymbol{ \tilde \eta}=B^{-\top}\boldsymbol \eta$.
   \STATE Let $\bold y = P \bold y'$, where $P$ is the permutation matrix that sorts $\bold y$ in increasing order.
   \STATE With probability $\alpha(\bold y, \boldsymbol{ \tilde \eta}, P)$, accept $\bold y$ and return $\bold x = B^{-1} \bold y$. Otherwise, go back to step 1.
\end{algorithmic}
\end{algorithm}

\section{Experiments}\label{sec:experiments}

\subsection{Simulation study} \label{sec:dirichlet_bias}

We begin our experiments with a simulation study that illustrates the biases incurred from fitting Dirichlet distributions to compositional data. 
Our procedure is as follows: 
\begin{enumerate}
\itemsep-0em
	\item Fix a ground-truth distribution $\bold x \sim p(\bold x)$ on the simplex. This will either be a Dirichlet where the (known) parameter value $ \text{\boldmath$ \alpha$}$ is drawn from independent $\text{Exp}(1)$ variates, or a CC where the (known) parameter value $ \text{\boldmath$ \lambda$}$ is sampled uniformly on the simplex. \label{step:prior}
	\item Fix a sample size $n$. This will range from $n=2$ to $n=20$.
	\item In each of one million trials, draw a sample of $n$ i.i.d. observations $\bold x_i \sim p(\bold x)$.
	\item Use the samples to compute one million MLEs for the mean parameter, under both a Dirichlet and a CC likelihood.
	\item Average the one million MLEs and subtract the true mean $\mathbb E_{p(\bold x)}[\bold x]$ to obtain estimates of the `empirical bias'.
	\item Repeat the steps for different values of $ \text{\boldmath$ \alpha$}$, $ \text{\boldmath$ \lambda$}$, and $n$, as prescribed.
\end{enumerate}

\begin{figure}
\centering
  \includegraphics[width=.95\linewidth]{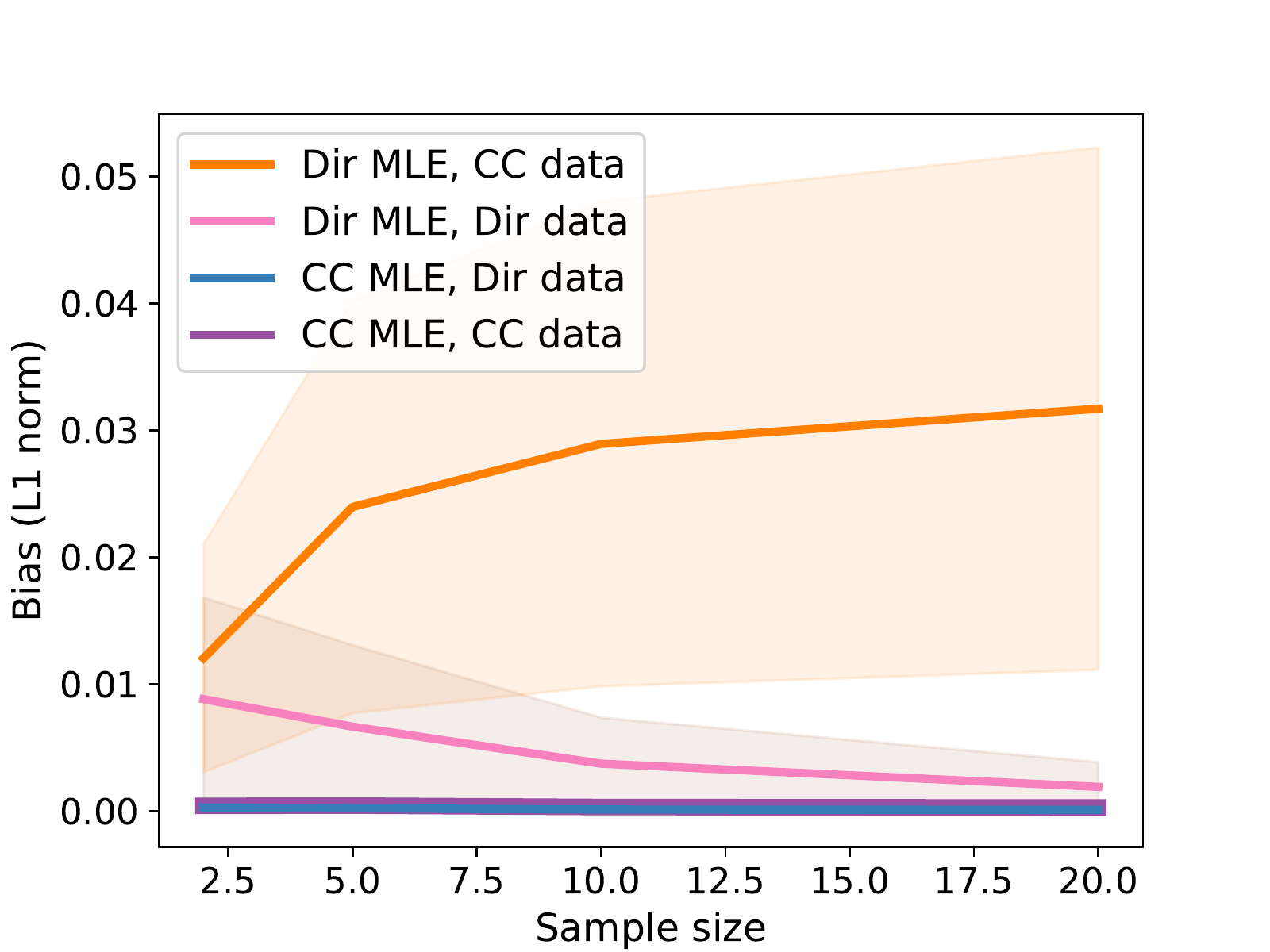}
  \caption{Empirical bias of the Dirichlet and CC MLEs as a function of sample size, for $K=3$ (other values of $K$ behaved similarly). The error bars show $\pm1$ standard deviation over different draws of the parameters of the ground-truth model from their respective priors. Regardless of whether the synthetic data is generated from a Dirichlet or a CC, only the CC estimator is unbiased.}
   \label{fig:dirichlet_bias}
\end{figure}

The results of this experiment are shown in figure \ref{fig:dirichlet_bias}.
As we already knew from the theory of exponential families, only the CC estimator is unbiased.
The Dirichlet estimator is, at best, asymptotically unbiased, and only in the unrealistic setting of a true Dirichlet generative process (pink line).
It is worth emphasizing that, even in this most-favorable case for the Dirichlet, the CC outperforms (pink line versus blue line).
The error bars show $\pm1$ standard deviation over different draws from the prior distribution of step \ref{step:prior}.
The large error bars on the Dirichlet, especially under non-Dirichlet data (orange line) indicate that its bias is highly sensitive to the true mean of the distribution, reaching up to several percentage points (relative to the unit-sum total), whereas the CC is unbiased across the board.
Lastly, note that while a sample size of 20 may seem small in the context of machine learning, this is in fact reasonable in terms of the ratio of observations to parameters, as larger samples typically call for more flexible models.

\subsection{UK election data} \label{sec:election}

\begin{figure}
\centering
  \includegraphics[width=.95\linewidth]{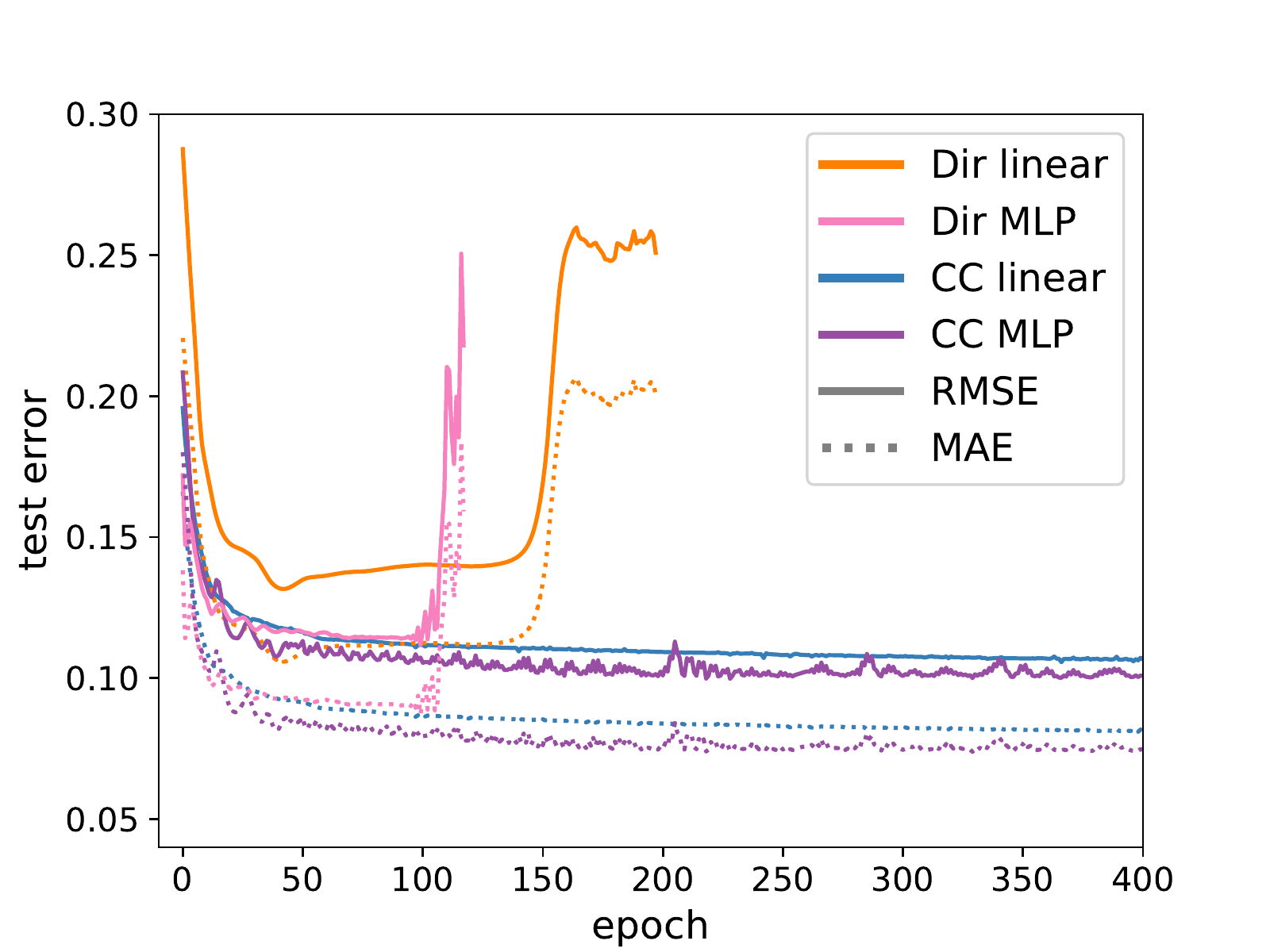}
  \caption{Test error for the linear and MLP models with a Dirichlet and CC log-likelihood. The CC objective is better-behaved, training more smoothly and converging to the MLE, unlike the Dirichlet log-likelihood which diverges, causing overfitting.}
   \label{fig:test_err}
\end{figure}

We next consider a real-world example from the 2019 UK general election \cite{uberoi2019general}.
The UK electorate is divided into 650 constituencies, each of which elects 1 member of parliament (MP) in a winner-takes-all vote.
Typically, each of the major parties will be represented in each constituency, and will win a share of that constituency's electorate.
Thus, our data consists of 650 observations\footnote{We split the data into an 80/20 training and test set at random.}, each of which is a vector of proportions over the four major parties (plus a fifth proportion for a `remainder' category, which groups together smaller parties and independent candidates).
We regress this outcome on four predictors: the region of the constituency (a categorical variable with 4 levels), an urban/rural indicator, the size of the constituency's electorate, and the voter turnout percentage from the previous general election.
While there is a host of demographic data that could be used to construct increasingly more informative predictors here, we stick to the reduced number of variables that were available in our original dataset, as the goal of our analysis is \emph{not} to build a strong predictive model, but rather to illustrate the advantages and disadvantages of using the novel CC distribution as opposed to the Dirichlet.
For the benefit of the latter, since our data contains zeros (not all major parties are represented in all constituencies), we add an insignificant $0.1\%$ share to all observations and re-normalize prior to modeling (without this data distortion, the Dirichlet would fail even more grievously).

We fit regression networks to this data with two different loss functions, one where we assume the response follows a Dirichlet likelihood \cite{campbell1987multivariate, hijazi2003analysis, hijazi2009modelling}, the other with our own CC likelihood instead.
We first fit the simplest linear version: for the Dirichlet, we map the inputs into the space of  $ \text{\boldmath$\alpha$} $ by applying a single single linear transformation followed by an $\exp(\cdot)$ activation function; for the CC, a single linear transformation is sufficient to map the inputs to the unconstrained space of $ \text{\boldmath$\eta$} $ (in statistical terminology, this corresponds to a generalized linear model with canonical link).
Secondly, we extend our linear model to a more flexible neural network by adding a hidden layer with 20 units and ReLU activations.
We train both models using Adam \cite{kingma2014adam}.

\begin{table}[t]
\caption{Test errors and runtime for our regression models of the UK election data. Both in the linear and MLP case, the CC model beats the Dirichlet counterpart.} \label{tab:election}
\vskip 0.15in
\begin{center}
\begin{small}
\begin{sc}
\begin{tabular}{lcccr}
\toprule
  & MAE & RMSE & ms/epoch\\
\midrule
Dir linear    & 0.105 & 0.132 & 4 \\
CC linear	& \textbf{0.075} & \textbf{0.101} & 8 \\
\hdashline[0.5pt/5pt]
Dir MLP    & 0.087 & 0.112 & 20 \\
CC MLP	& \textbf{0.072} & \textbf{0.097} & 35 \\
\bottomrule
\end{tabular}
\end{sc}
\end{small}
\end{center}
\vskip -0.1in
\end{table}

\begin{figure}
\centering
    \includegraphics[width=.95\linewidth]{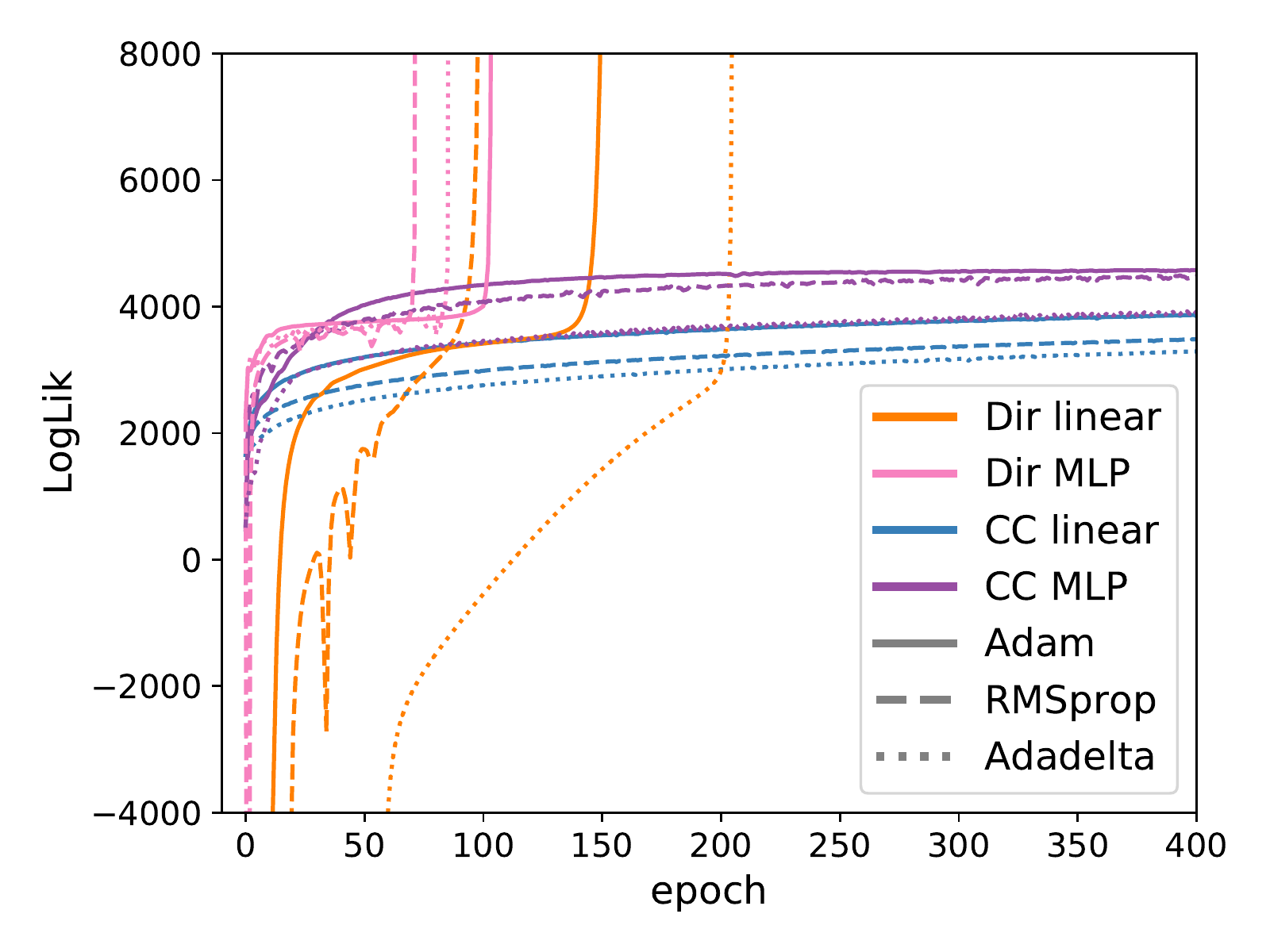}
  \caption{Log-likelihood during training with different optimizers for the Dirichlet and the CC models. Consistently across optimizers, the CC objective converges to the MLE and the Dirichlet diverges.}
   \label{fig:optimizers}
\end{figure}

\begin{figure}
\centering
  \includegraphics[width=.95\linewidth]{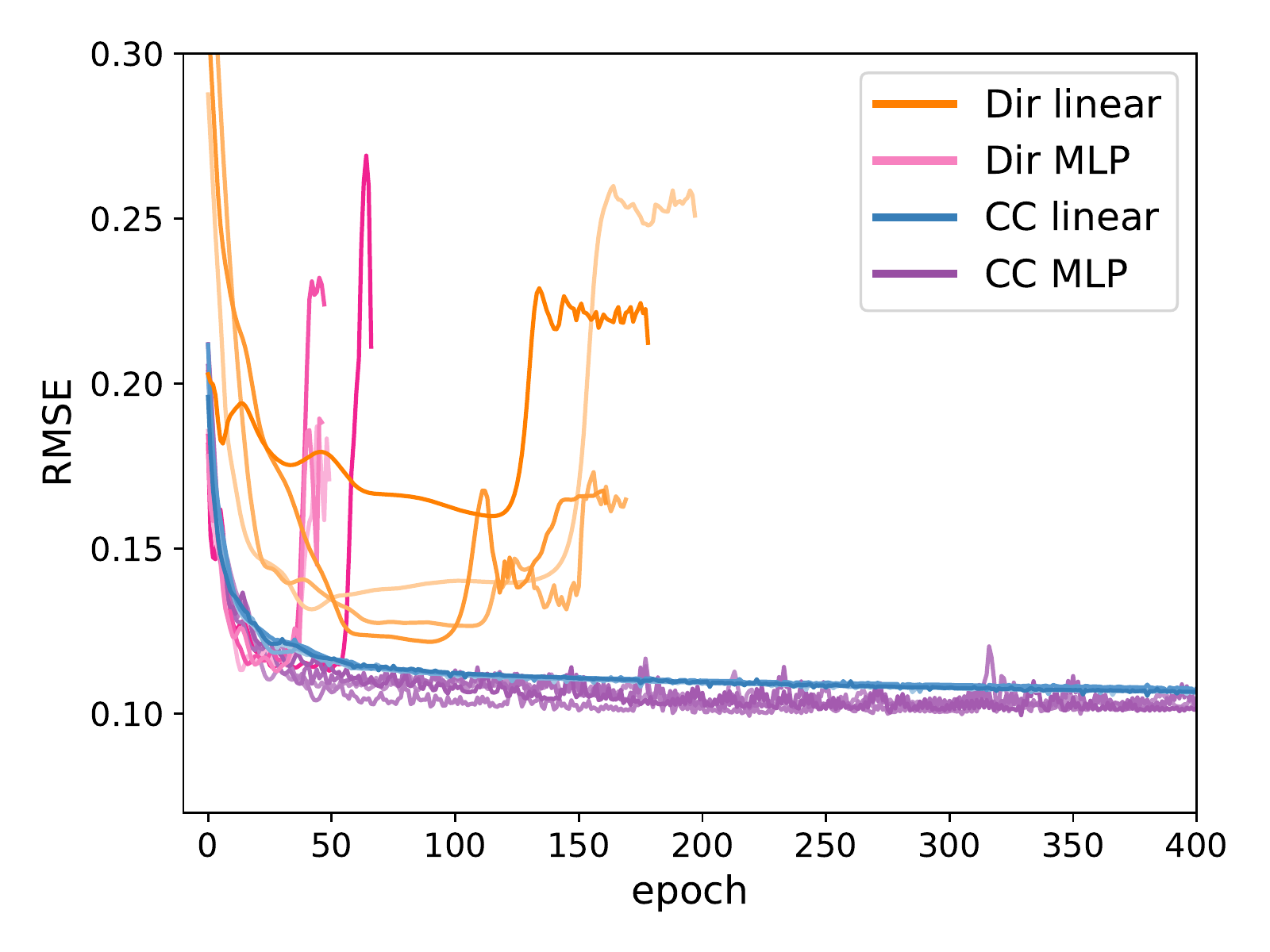}
  \caption{Test error for different random initializations of the model parameters. The CC model converges, whereas the Dirichlet diverges along a highly variable path in the parameter space, and the models it finds in this path are suboptimal.}
  \label{fig:seeds_election}
\end{figure}

The CC models achieve better test error, with gains ranging between $10$\% and $30$\% in the $L_1$ and $L_2$ sense, as shown in table \ref{tab:election} and figure \ref{fig:test_err}.\footnote{The $L_1$ and $L_2$ metrics are used for illustration purposes. One could, of course, optimize these metrics directly, however this would either jeopardize the probabilistic interpretation and statistical rigor of the model, or require the addition of an intractable normalizing constant to the log-likelihood.} 
Moreover, the CC reliably converges to a performant model (which in the linear case corresponds to the MLE), whereas the Dirichlet likelihood diverges along a highly variable path in the parameter space that correspond to suboptimal models.
We verify also that this behavior happens irrespective of the optimizer used to train the model (figure \ref{fig:optimizers}), and is consistent across random initializations of the model parameters (figure \ref{fig:seeds_election}).
Note that the Dirichlet MLP diverges faster than its linear counterpart, likely because the increased flexibility afforded by the MLP architecture allows it to place a spike on a training point more quickly.

%
%

\begin{figure}
\centering
  \includegraphics[width=.95\linewidth]{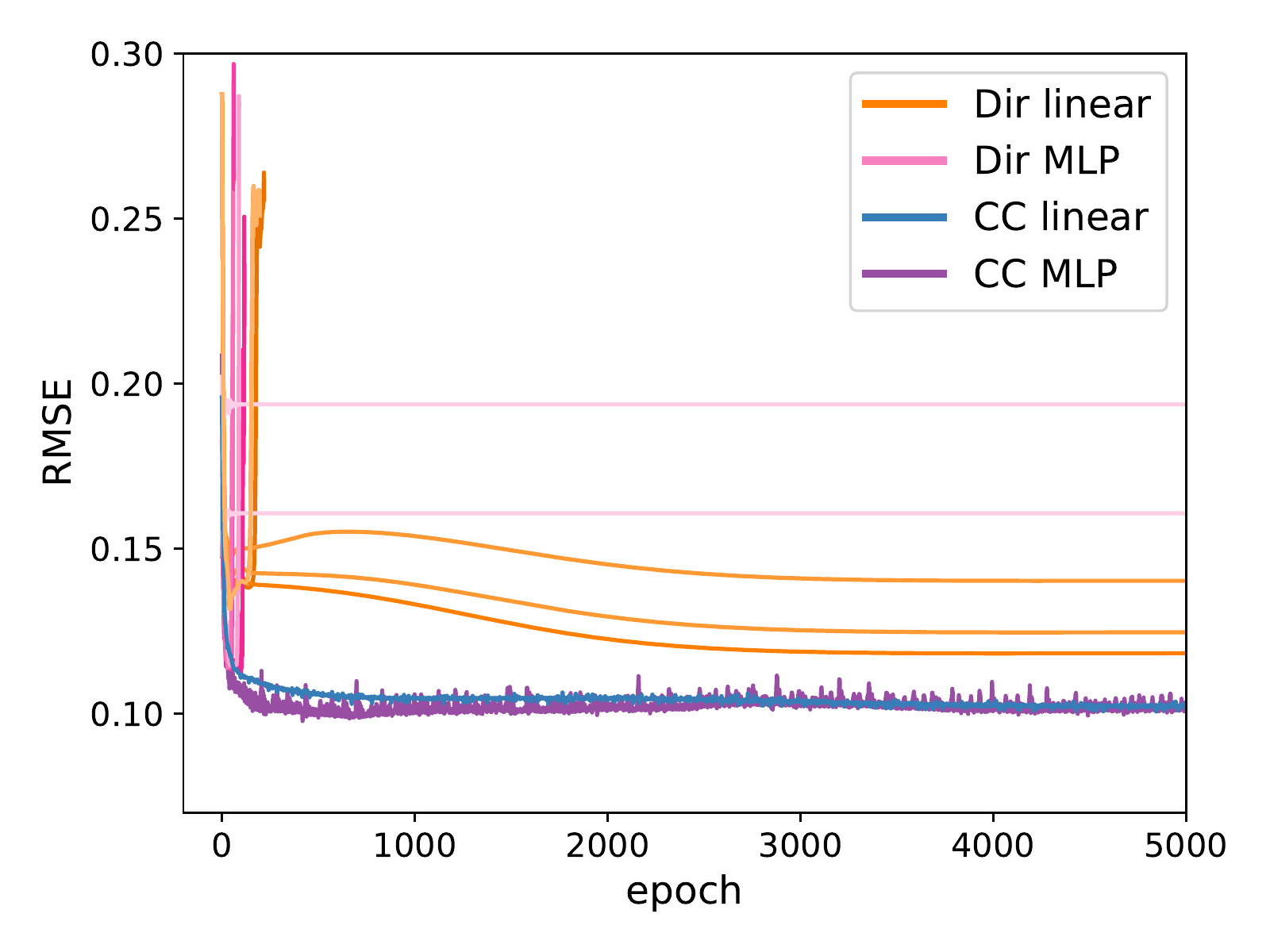}
  \caption{Test error for different regularization strengths for the Dirichlet models. Lighter shades of pink and orange indicate increasing regularization strengths. While sufficiently strong regularization does stabilize the Dirichlet, preventing it from diverging, it still does not outperform the (unregularized) CC.}
   \label{fig:reg_election}
\end{figure}

Naturally, one might ask whether the unstable behavior of the Dirichlet can be fixed through regularization, as a suitable penalty term might be able to counterbalance the detrimental flexibility of the distribution.
We test this hypothesis empirically by adding an $L_2$ penalty (applied to the weights and biases of the regression network) to the objective, with a varying coefficient to control the regularization strength.
The results are shown in figure \ref{fig:reg_election}; while strong regularization does stabilize the Dirichlet, it is still unable to outperform the (unregularized) CC and it is slower to converge.

In terms of runtime, we find that the computational cost of the CC and Dirichlet gradients are of the same order of magnitude (table \ref{tab:election}).
While the CC models are slower per gradient step, they are able to find better regression functions in fewer steps.
What's more, the comparison is somewhat unfair to the CC, as the Dirichlet gradients exploit specialized numerical techniques for evaluating the digamma function that have been years in the making, whereas the same is not yet true for our novel distribution.

\subsection{Model compression}

Our last experiment considers a typical model compression task \citep{bucilua2006model,ba2014deep,hinton2015distilling}, where we have access to an accurate `teacher' model that is expensive to evaluate, which we use in order to train a cheaper `student' model, typically a neural network with few layers.
By training the student network to predict the fitted values of the teacher model, it can achieve better generalization than when trained on the original labels, as was shown by \citet{bucilua2006model,ba2014deep,hinton2015distilling}.
These fitted values provide `soft targets' that are typically located close to the extrema of the simplex, though we can also bring them towards the centroid while conserving their relative order by varying the temperature, $T$, of the final softmax \cite{hinton2015distilling}.

We build on the MNIST experiment of \citet{hinton2015distilling}; we first train a teacher neural net with two hidden layers of 1200 units and ReLU activations, regularized using batch normalization, 
and we use its fitted values to train smaller student networks with a single hidden layer of 30 units.
We fit the student networks to the soft targets using the categorical cross-entropy (XE) loss, as well as a Dirichlet (as per \citet{sadowski2018neural}) and a CC log-likelihood.
The latter is equivalent to adding the appropriate normalization constant to the XE, thus giving a correct probabilistic interpretation to the procedure.
Note that the XE and the CC objective result in the same estimator at optimality: the empirical mean of the data (we know this is true theoretically from section \ref{sec:exp_fam}).
However, the two objectives define different optimization landscapes, so the question of which one to use becomes an empirical one.

\begin{figure}
\centering
  \includegraphics[width=.95\linewidth]{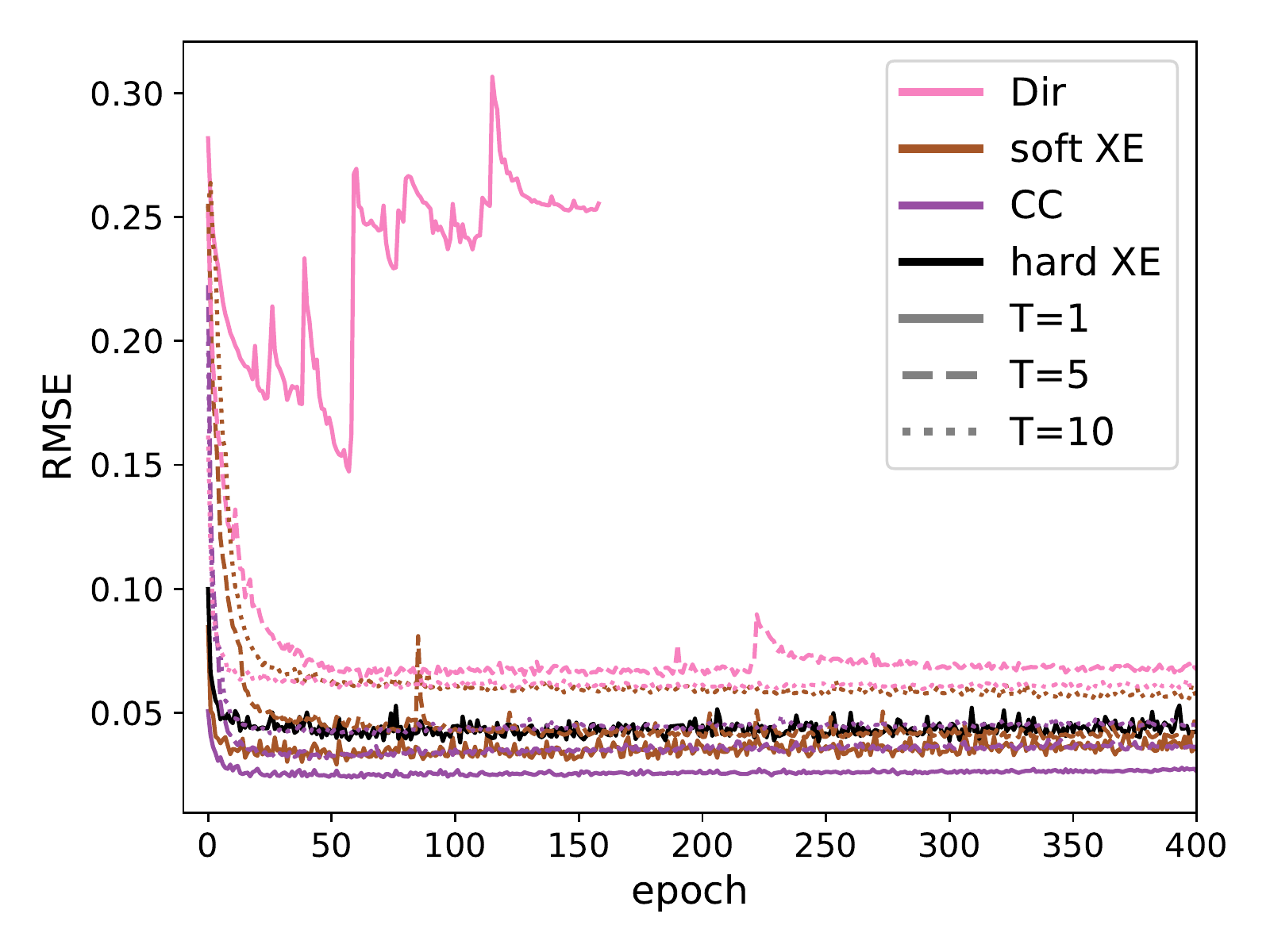}
  \caption{Test error at different temperatures for the Dirichlet, XE and CC objectives. For each student model, this error measures the $L_2$ difference between its fitted values and those of the teacher model, on the test set.}
   \label{fig:student_err}
\end{figure}

In this case, we found that using the CC objective leads to a lower misclassification rate on the test set than the XE (table \ref{tab:compression}), as well as a lower RMSE (measured against the soft targets outputted by the trained teacher model when evaluated on the test set), and did so consistently across temperatures (figure \ref{fig:student_err}).
Both the CC and XE objectives worked substantially better than the Dirichlet likelihood, and better than training the same architecture on the original hard labels.
Note also that the Dirichlet performs very poorly and is unstable in the unadjusted temperature setting  ($T=1$), as the soft targets are very extremal and lead to numerical overflow, but performs much better at high temperatures.


%

\begin{table}[t]
\caption{Test errors for the student models trained under the Dirichlet, XE, and CC objectives, as well as using the hard labels instead of the teacher model. We show the best values obtained over 5 random initializations and the 3 temperature settings of figure \ref{fig:student_err}.} \label{tab:compression}
\vskip 0.15in
\begin{center}
\begin{small}
\begin{sc}
\begin{tabular}{lcccr}
\toprule
 Objective & Accuracy  & RMSE & s/epoch\\
\midrule
Dirichlet 	& 90.6\% & 0.041 & 0.9 \\
Soft XE    & 94.9\% & 0.029 & 0.8 \\
CC    & \textbf{95.6}\% & \textbf{0.024} & 2.2 \\
\hdashline[0.5pt/5pt]
Hard XE    & 93.2\% & -- & 0.8 \\
\bottomrule
\end{tabular}
\end{sc}
\end{small}
\end{center}
\vskip -0.1in
\end{table}

\subsection{Latent variable models}

While we have demonstrated empirical improvements from using the CC to model compositional outcomes, so far we have not found similar gains when using the CC to model latent variables, for example, in mixed-membership models such as LDA \cite{blei2003latent, pritchard2000inference}.
Training a `Latent-CC-Allocation' topic model on a corpus of $\sim$2,000 NeurIPS papers gave similar learned topics and held-out perplexities as vanilla LDA (562 vs 557 per word, respectively).
However, while the Dirichlet is the conjugate prior for the multinomial, a CC prior with a multinomial likelihood leads to an intractable posterior, complicating optimization (which, in this case, is enabled by means of our sampling and reparameterization algorithms from appendix C).
We note also that the CC does not share the sparsity-inducing properties of the Dirichlet, nor can it approximate the categorical distribution arbitrarily well, unlike other continuous relaxations that have been proposed in the literature \cite{jang2016categorical, maddison2016concrete, potapczynski2019invertible}.
Nevertheless, given the widespread use of simplex-valued latent variables in probabilistic generative models, the use of the CC distribution in this context remains an open question.

\section{Conclusion}\label{sec:conclusions}

Our results demonstrate the theoretical and empirical benefits of the CC, which should hold across a range of applied modeling contexts.
To conclude, we summarize the main contributions of our work:
\begin{itemize}
\itemsep0em
	\item We have introduced the CC distribution, a novel exponential family defined on the simplex, that resolves a number of long-standing limitations of previous models of compositional data. 
	\item We have fully characterized our distribution and discussed its theoretical properties. Of particular importance is the sufficient statistic, which guarantees unbiased estimators, and the favorable behavior of the log-likelihood, which leads to ease of optimization and robustness to extreme values.
	\item Empirically, the CC defines probabilistic models that outperform their Dirichlet counterparts and other competitors, and allows for a rich class of regression functions, including neural networks.
	\item We have also designed and implemented novel and efficient rejection sampling algorithms for the CC.
\end{itemize}

Taken together, these findings indicate the CC is a valuable density for probabilistic machine learning with simplex-valued data.

\section*{Acknowledgements}

We thank Andres Potapczynski, Yixin Wang and the anonymous reviewers for helpful conversations, and the
Simons Foundation, Sloan Foundation, McKnight Endowment Fund,
NSF 1707398, and the Gatsby Charitable Foundation for support.

\bibliography{ref}
\bibliographystyle{icml2020}

\end{document}


\icmltitle{Supplementary material for: \\The continuous categorical: a novel simplex-valued exponential family}

\vskip 0.3in




\section{Derivation of the normalizing constant} \label{sec:norm_const}

For clarity, we start by recalling the expression that we aim to show (equation 7 from the main text), and we make the dependence on $K$ explicit by writing  $C_K(\text{\boldmath$\eta$})$:
\begin{equation} \label{eq:norm_const2} 
 C_K(\text{\boldmath$\eta$}) =
\left(
(-1)^{K+1} \sum_{k=1}^K 
\frac
 { { \exp\left(\eta_k\right)} }
 { \prod_{i \ne k } \left(\eta_i - \eta_k\right)}
 \right)^{-1}
 .
\end{equation}

%

\textbf{Proof:} we proceed by induction on $K$. The base case $K=2$ can be integrated directly:
\begin{align} \nonumber
C_2(\boldsymbol \eta) &= 
\left(
\int_0^1 \exp( \eta_1 x_1) dx_1
\right) ^{-1}
\\ \nonumber
&=  
\left(
\frac{e^{\eta_1} - 1}{\eta_1}
\right) ^{-1}
\\
&=
\left(
- \frac{e^{\eta_1}}{\eta_2 - \eta_1} - \frac{ e^{\eta_2}}{\eta_1 - \eta_2}
\right) ^{-1},
\end{align}
where the last equality follows from $\eta_2=0$.

For the inductive step, we assume that equation \ref{eq:norm_const2} gives the correct normalizing constant for $K-1$, and compute the integral for $K$:
\begin{align} \nonumber
C_K(\text{\boldmath$\eta$})^{-1}
 &= \int_{{\mathbb{S}^{K-1}}} \exp(\text{\boldmath$\eta$}^\top \bold x) d\mu
\\
&= \int_0^1 \int_0^{1-x_1} \cdots \int_0^{1-x_1-\cdots-x_{K-2}} \exp\left(\sum_{i=1}^{K-1} \eta_i x_i\right)
dx_{K-1} \cdots dx_2 dx_1.
\end{align}
For the innermost integral, we have:
\begin{align} \nonumber
\int_0^{1-x_1-\cdots-x_{K-2}} &\exp\left(\sum_{i=1}^{K-1} \eta_i x_i\right)
dx_{K-1} 
\\ \nonumber
&=
\exp\left(\sum_{i=1}^{K-2} \eta_i x_i\right) \int_0^{1-x_1-\cdots-x_{K-2}} 
\exp(\eta_{K-1} x_{K-1}) dx_{K-1} 
\\ \nonumber
&= 
\exp\left(\sum_{i=1}^{K-2} \eta_i x_i\right) \left[ \frac{1}{\eta_{K-1}} \exp(\eta_{K-1}t) \right]_{t=0}^{t=1-x_1-\cdots-x_{K-2}} 
\\ \nonumber
&= 
 \frac{1}{\eta_{K-1}} 
\exp\left(\sum_{i=1}^{K-2} \eta_i x_i\right)   \left[\exp(\eta_{K-1}(1-x_1-\cdots-x_{K-2})) - 1 \right]
\\
 &= \label{eq:inner}
 \frac{1}{(\eta_{K-1}- \eta_K)} 
 \left[
\exp(\eta_{K-1}) \exp\left(\sum_{i=1}^{K-2} (\eta_i - \eta_{K-1}) x_i\right)
 -
 \exp\left(\sum_{i=1}^{K-2} \eta_i x_i\right)
 \right]
.
\end{align}
Letting $\eta^{(1)}_i = \eta_i - \eta_{K-1}$ for $i = 1,\dots,K-1$, by inductive hypothesis we have that:
\begin{align} \nonumber
 C_{K-1}(\text{\boldmath$\eta$}^{(1)})^{-1}
&=
\int_0^1 \int_0^{1-x_1} \cdots \int_0^{1-x_1-\cdots-x_{K-3}} 
 \exp\left(\sum_{i=1}^{K-2} (\eta_i - \eta_{K-1}) x_i\right)
dx_{K-2} \cdots dx_2 dx_1 
\\ \nonumber
&=
(-1)^{K} \sum_{k=1}^{K-1} 
 \frac
 { { \exp\left(\eta_k^{(1)}\right)} }
 { \prod_{i \ne k } \left(\eta_i^{(1)} - \eta_k^{(1)}\right)}
\\ \label{eq:piece1}
&=
(-1)^{K} \sum_{k=1}^{K-1} 
 \frac
 { { \exp\left(\eta_i - \eta_{K-1}\right)} }
 { \prod_{i \ne k } \left(\eta_i - \eta_k\right)}
 .
\end{align}
Similarly, letting $\eta^{(2)}_i = \eta_i$  for $i = 1,\dots,K-2$, and $\eta^{(2)}_{K-1} = 0$, we have that:
\begin{align} \nonumber
 C_{K-1}(\text{\boldmath$\eta$}^{(2)})^{-1}
&=
\int_0^1 \int_0^{1-x_1} \cdots \int_0^{1-x_1-\cdots-x_{K-3}} 
 \exp\left(\sum_{i=1}^{K-2} \eta_i x_i\right)
dx_{K-2} \cdots dx_2 dx_1 
\\ \nonumber
&=
(-1)^{K} \sum_{k=1}^{K-1} 
 \frac
 { { \exp\left(\eta_k^{(2)}\right)} }
 { \prod_{i \ne k } \left(\eta_i^{(2)} - \eta_k^{(2)}\right)}
\\ \nonumber
&=
(-1)^{K} \left[ \sum_{k=1}^{K-2} 
 \frac
 { { \exp\left(\eta_k \right)} }
 { (- \eta_k) \prod_{i \ne k } \left(\eta_i - \eta_k\right)}
 + \frac{1}{\prod_{i=1}^{K-2} \eta_i}
\right] 
\\
&= \label{eq:piece2}
(-1)^{K} \left[ \sum_{k=1}^{K-2} 
 \frac
 { { \exp\left(\eta_k \right)} }
 { - ( \eta_k - \eta_K) \prod_{i \ne k } \left(\eta_i - \eta_k\right)}
 + \frac{\exp(\eta_K)}{\prod_{i=1}^{K-2} (\eta_i - \eta_K)}
\right] 
.
\end{align}
Plugging (\ref{eq:piece1}) and (\ref{eq:piece2}) back into (\ref{eq:inner}), we find:
\begin{align*}
 C_K(\text{\boldmath$\eta$})^{-1} =
(-1)^{K+1} \sum_{k=1}^K 
R_k(\text{\boldmath$\eta$}) { \exp\left(\eta_k\right)} 
,
\end{align*}
where the coefficients $R_k(\text{\boldmath$\eta$})$ gather the terms that multiply each $\exp(\eta_k)$ term.
For $k=1,\dots,K-2$, both (\ref{eq:piece1}) and (\ref{eq:piece2}) contribute to the coefficient:
\begin{align} \nonumber
R_k(\text{\boldmath$\eta$}) 
&= 
\frac{1}{\eta_{K-1} - \eta_{K}}
\left[
- \frac{1}{\prod_{1 \le i \le K-1, i \ne k} (\eta_i - \eta_k)}
+ \frac{1}{\prod_{1 \le i \le K, i \ne k, i \ne K-1} (\eta_i - \eta_k)}
\right]
\\ \nonumber
&= 
\frac{1}{\eta_{K-1} - \eta_{K}}
\left[
\frac{- \eta_{K} + \eta_k + \eta_{K-1} - \eta_k}{\prod_{1 \le i \le K, i \ne k} (\eta_i - \eta_k)}
\right]
\\
&=
\frac{1}{\prod_{i \ne k} (\eta_i - \eta_k)}
.
\end{align}
The $(K-1)$th coefficient can be computed more easily as it only appears in (\ref{eq:piece1}):
\begin{align} \nonumber
R_{K-1}(\text{\boldmath$\eta$}) &=
- \frac{1}{(\eta_{K-1} - \eta_K)} \frac{1}{\prod_{1 \le i \le K-2} (\eta_i - \eta_{K-1})}
\\
&=
\frac{1}{\prod_{i \ne K-1} (\eta_i - \eta_{K-1})},
\end{align}
and similarly, the $K$th coefficient appears only in (\ref{eq:piece2}):
\begin{align} \nonumber
R_{K}(\text{\boldmath$\eta$}) &=
\frac{1}{(\eta_{K-1} - \eta_K)} \frac{1}{\prod_{1 \le i \le K-2} (\eta_i - \eta_{K-1})}
\\
&= 
\frac{1}{\prod_{i \ne K} (\eta_i - \eta_{K})}. 
\end{align}

This completes the proof. $\square$

\textbf{Remark.} For completeness, we also include the normalizing constant written in terms of the parameterization of the original density in equation 2 of the main text:
\begin{equation*} 
\int_{\mathbb S^{K-1}} \prod_{i=1}^K \lambda_i^{x_i} d \mu(\bold x) = 
(-1)^{K+1} \sum_{k=1}^K \frac
 { \lambda_k }
 {  \prod_{i\ne k} \log{\frac{\lambda_{i}}{\lambda_{k}}} }
.
\end{equation*}

\section{Additional properties of the CC distribution} \label{sec:additional_properties}

\subsection{Mean and covariance}

As mentioned in the main manuscript (section 3.5), by standard properties of exponential families, the mean and covariance of the CC can be obtained by differentiating the normalizing constant. For completeness, we include these results here. If $\bold x \sim \mathcal{CC}(\text{\boldmath$\eta$})$, then the mean of $\bold x$ is given by:
\begin{equation}
\mathbb E [x_i] = - \frac{\partial}{\partial \eta_i} \log {C}(\text{\boldmath$\eta$}),
\end{equation}
and the covariance is given by:
\begin{equation}
\text{cov}(x_i, x_j) = - \dfrac{\partial^2}{\partial \eta_i \partial \eta_j} \log  {C}(\text{\boldmath$\eta$}).
\end{equation}

\subsection{KL Divergence}

The KL divergence between two CC variates can be computed directly from their means: 
\begin{align} \nonumber
 KL(p(\bold x| \text{\boldmath$  \eta$})||p(\bold x| \text{\boldmath$ \tilde \eta$}) )
&= \mathbb E_{p(\bold x |  \text{\boldmath$  \eta$})} \left[ \log \frac{p(\bold x |  \text{\boldmath$  \eta$})}{p(\bold x |  \text{\boldmath$ \tilde \eta$})} \right]
\\ \nonumber
&= \mathbb E_{ p(\bold x |  \text{\boldmath$  \eta$})} \left[  \log C(\text{\boldmath$\eta$}) - \log C(\text{\boldmath$\tilde \eta$})  + \sum_{i=1}^{K-1} (\eta_i - \tilde \eta_i) x_i \right]
\\
&=  \log C(\text{\boldmath$\eta$}) - \log C(\text{\boldmath$\tilde \eta$}) + (\boldsymbol \eta -  \boldsymbol{ \tilde \eta} )^\top \mathbb E_{p(\bold x |  \text{\boldmath$  \eta$})} [ \bold x ] 
 .
\end{align}

\subsection{Moment generating function}

The moment generating function of the CC distribution can be written directly in terms of the normalizing constant:
\begin{align} \nonumber
M_{\bold x}(\bold t) &= \mathbb E [ e^{\bold t^\top \bold x} ]
\\ \nonumber
&= \int_{\mathbb{S}^{K-1}} e^{\bold t^\top \bold x}   C(\text{\boldmath$\eta$}) e^{\text{\boldmath$\eta$}^\top \bold x} d\mu
\\ \nonumber
&=  C(\text{\boldmath$\eta$}) \int_{\mathbb{S}^{K-1}} e^{(\bold t + \text{\boldmath$\eta$})^\top \bold x}  d\mu
\\
&= \frac{ C(\text{\boldmath$\eta$})  }{ C( \bold t + \text{\boldmath$\eta$})  }.
\end{align}
The characteristic function can be derived similarly.

\subsection{Marginalization} \label{sec:marginalization}

Unlike the Dirichlet, the CC is not preserved under marginalization, even when allowing transformations of the parameter vector. In other words, if $(x_1, \dots, x_{K-1}) \sim \mathcal{CC}(\eta_1,\dots,\eta_{K-1})$, then it is not true that $x_1 \sim \mathcal{CC} (\eta_1)$, nor that $(x_1,\dots,x_{K-2}) \sim \mathcal{CC}(\eta_1,\dots,\eta_{K-2})$. It is not even true that $x_1 \sim \mathcal{CC}(\tilde \eta_1)$, nor that $(x_1,\dots,x_{K-2}) \sim \mathcal{CC}(\tilde \eta_1,\dots,\tilde \eta_{K-1})$, for any $ \text{\boldmath$ \tilde \eta$}$. This can be seen easily by integrating out the case $K=3$:
\begin{align} \nonumber
\int_{0}^{1-x_1} C(\eta_1, \eta_2) \exp(\eta_1 x_1 + \eta_2 x_2) dx_2 
&= C(\eta_1, \eta_2)  \exp(\eta_1 x_1) 
\left[ \frac{ \exp(\eta_2 t) } {\eta_2}  \right]_{t=0}^{t=1-x_1}
\\
&=
\frac{C(\eta_1, \eta_2) \exp(\eta_2)}{\eta_2}  \exp( (\eta_1 - \eta_2) x_1) 
- \frac{C(\eta_1, \eta_2)}{\eta_2}  \exp( \eta_1 x_1) ,
\end{align}
which is not of the form $C(\tilde \eta_1) \exp(\tilde \eta_1 x_1)$ for any $\tilde \eta_1$.


As a direct consequence, we cannot use a stick-breaking construction \citep{connor1969concepts, paisley2010stick} to simulate CC variates from 1-dimensional CB variates, as with the Dirichlet and the Beta distributions.


\section{Sampling}\label{sec:samplers_supp}

In this section, we develop sampling algorithms for the CC distribution and analyze their performance empirically. 
We also describe how to use our samplers to obtain reparameterization gradients \cite{kingma2013auto}.

\subsection{The `naive' rejection sampler} \label{sec:naive_sampler}

Given the form of the CC density function, a rejection sampling scheme follows readily by combining independent 1-dimensional CB draws (algorithm \ref{alg:naive_sampling}).
\begin{algorithm}
   \caption{Naive sampler}
   \label{alg:naive_sampling}
   {\bfseries Input:} target distribution $\mathcal{CC}(\text{\boldmath$  \lambda$})$.
   \\
   {\bfseries Output:} sample $\bold x$ drawn from target.
   \\ \vspace{-4mm}
   \begin{algorithmic}[1]
   \STATE For $i=1,\dots,K-1$, draw $x_i \sim \mathcal{CC}(\lambda_i, \lambda_K)$ independently.
   \STATE If $\sum_{i=1}^{K-1} x_i > 1$, go back to step 1, otherwise return $\bold x = (x_1, \dots, x_{K-1})$.
\end{algorithmic}
\end{algorithm}

To see why algorithm \ref{alg:naive_sampling} achieves the desired distribution, firstly note that by independence, the distribution produced in step 1 is:
\begin{align} 
p_{\text{step}1}(\bold x) 
\propto \prod_{i=1}^{K-1} \lambda_i^{x_i} \lambda_K^{1-x_i} \propto \lambda_1^{x_1} \cdots \lambda_{K-1}^{x_{K-1}} \lambda_K^{1-x_1-\cdots-x_{K-1}}. 
\end{align}
This is precisely the density we seek, except it is drawn on $[0,1]^{K-1}$ instead of the simplex. 
Step 2 rejects all samples that fall outside the simplex, thus achieving the target distribution.

The obvious shortcoming of this sampling approach is that, even for moderate values of $K$, the proportion of rejections becomes large. 
This is particularly troublesome in the balanced case, $\bold x \sim \mathcal{CC}(1/K,\dots,1/K)$, which is equivalent to drawing uniformly on $[0,1]^{K-1}$ and rejecting whenever we fall outside of a simplex of measure $1/(K-1)!$.
In other words, we accept with a probability that decays factorially in dimension. 

\subsubsection{Reparameterization} \label{sec:reparam}

The 1-dimensional CB distribution can be reparameterized using the analytical expression for the inverse CDF, derived by \citet{loaiza2019continuous}.
In this section we extend the strategy to a multivariate analogue for the CC distribution.
The underlying idea is that the rejection step in algorithm \ref{alg:naive_sampling} only depends on the  $L_1$ norm of the proposal, but not on the parameter.
This implies that, once we find an accepted proposal, we can use the inverse CDF reparameterization directly, without requiring a correction term as per the general framework for acceptance-rejection reparameterization gradients \cite{naesseth2016reparameterization}.

Our aim is to write $\bold x = g(\bold u, \text{\boldmath$  \lambda$})$, where the density of $\bold u $ does not depend on $\text{\boldmath$  \lambda$}$.
To this end, write $F(x|\lambda_i,\lambda_K)$ for the CDF of $x \sim \mathcal{CC}(\lambda_i, \lambda_K)$. 
Note that this expression will follow readily from an equivalent CB distribution, as $\mathcal{CC}(\lambda_i, \lambda_K) = \mathcal{CB}(\lambda_i/(\lambda_i + \lambda_K))$. 
For each $i = 1,\dots,K-1$, applying the inverse CDF component-wise on each of $u_i \overset{iid}{\sim} U(0,1)$ results in $F^{-1}(u_i|\lambda_i,\lambda_K) \sim \mathcal{CC}(\lambda_i, \lambda_K)$. 
Thus, the vector 
\begin{align}
& \bold F^{-1}( \bold u | \text{\boldmath$  \lambda$}) := 
 [F^{-1}(u_1|\lambda_1,\lambda_K), \dots , F^{-1}(u_{K-1} |\lambda_{K-1},\lambda_K)] 
\end{align} 
provides a differentiable reparameterization of the distribution $\mathcal{CC}(\text{\boldmath$  \lambda$})$, provided $\bold u$ was drawn from the pre-image of the simplex ${\mathbb{S}^{K-1}}$ under the mapping $\bold F^{-1}$, or in other words, provided that $\bold u \in \bold F({\mathbb{S}^{K-1}})$. The rejection step simply guarantees that we find a sample of uniforms inside this region, but once we have found such a sample, it will lie in the interior of the region with probability 1, and therefore we can differentiate through the transformation as desired:
\begin{align}
\frac{\partial \bold x}{\partial \text{\boldmath$  \lambda$}} = \frac{\partial }{\partial  \text{\boldmath$  \lambda$}}\bold F^{-1}( \bold u |  \text{\boldmath$  \lambda$}) .
\end{align}
We formalize this reparameterization in algorithm \ref{alg:reparam}.

\begin{algorithm*}
   \caption{Reparameterized rejection sampler}
   \label{alg:reparam}
   {\bfseries Input:} target distribution $\mathcal{CC}(\text{\boldmath$  \lambda$})$.
   \\
   {\bfseries Output:} a sample $\bold u$ such that $\bold F^{-1}( \bold u | \text{\boldmath$  \lambda$}) \sim \mathcal{CC}(\text{\boldmath$  \lambda$})$.
   \begin{algorithmic} [1]
   \STATE For $i=1,\dots,K-1$, draw $u_i \sim U(0,1)$ and set $x_i = F^{-1}(x|\lambda_i, \lambda_K)$.
   \STATE If $\sum_{i=1}^K x_i > 1$, return to step 1, otherwise return $\bold u = (u_1, \dots, u_{K-1})$.
\end{algorithmic}
\end{algorithm*}

\subsection{The ordered rejection sampler} \label{sec:ordered_sampler}


An analysis of algorithm \ref{alg:naive_sampling} reveals two relevant observations.
Firstly, the simulation of each $x_j \sim \mathcal{CC}(\lambda_j, \lambda_K)$ in step 1 involves computing the inverse cdf $F^{-1}(\cdot|\lambda_i, \lambda_K)$, which is more expensive than computing the cumulative sum $\sum_{i=1}^j x_i$ of the draws. 
It therefore pays to recompute the cumulative sums after each draw and go directly to the rejection step as soon as it exceeds 1.
Secondly, note that we do not generally expect the components of $\text{\boldmath$  \lambda$}$ to be balanced.
Thus, even though simulating each $x_i$ in step 1 requires the same amount of computation, those drawn from smaller values of $\lambda_i$ are more likely to be close to 0 than those drawn from higher values of $\lambda_i$.
The dimensions that are more likely to be close to 0 are also less likely to make our cumulative sum exceed the rejection threshold.
It therefore also pays to draw the $x_i$ components in order of decreasing $\lambda_i$.

These remarks motivate an improved sampling scheme, which we call the ordered rejection sampler (algorithm \ref{alg:naive+}).
Empirically, we find that this sampler substantially reduces the rejection rate (see figure \ref{fig:rejection_rates_hist}) as well as the computation time (by not only rejecting less, but also rejecting sooner).
However, this sampler performs poorly when $\text{\boldmath$  \lambda$}$ is balanced; such a setting leaves little room for improvement from the re-ordering operation, and the resulting sampler is similar to the naive rejection sampler.
This motivates a further sampling scheme that we introduce in the following section, but further improvements to this sampler are left to future work.
Lastly, we note that the reparameterization scheme of section \ref{sec:reparam} can be modified trivially to apply here also.


\begin{algorithm}
   \caption{Ordered rejection sampler}
   \label{alg:naive+}
   {\bfseries Input:} target distribution $\mathcal{CC}(\text{\boldmath$  \lambda$})$.
   \\
   {\bfseries Output:} sample $\bold x$ drawn from target.
   \\ \vspace{-4mm}
   \begin{algorithmic}[1]
   \STATE Find the permutation $\pi$ that orders $\text{\boldmath$  \lambda$}$ from largest to smallest, and let  $\text{\boldmath$\tilde  \lambda$} = \pi( \text{\boldmath$  \lambda$})$.
   \STATE Set the cumulative sum $c \leftarrow 0$ and $i \leftarrow 2$.
   \WHILE{$c<1$}
   \STATE Draw $u_i \sim U(0,1)$.
   \STATE Set $x_{i} = F^{-1}(u_i|\tilde\lambda_{i}, \tilde\lambda_1)$.
   \STATE Set $c \leftarrow c + x_{i}$.
   \STATE Set $i \leftarrow i+1$.
   \ENDWHILE
   \STATE If $c>1$, go back to step 2.   
   \STATE Set $x_1 = 1 - \sum_{i=2}^K x_i$.
   \STATE Return $\bold x = \pi^{-1}(x_1,\dots,x_K)$.
\end{algorithmic}
\end{algorithm}



\subsection{The permutation sampler}\label{sec:perm_sampler}

Next, we develop a permutation sampler that performs particularly well for configurations of $\text{\boldmath$  \lambda$}$ that are balanced (those that lead to distributions that are close to uniform).
Our key insights here are that the unit cube can be partitioned into simplexes, each of which corresponds to a permutation of its dimensions, and that the CC distribution is, in a sense, `invariant' over these permutations.

\subsubsection{Partitioning the cube into simplexes}

Let $\mathcal{R} = [0,1]^{K-1}$, the unit cube. 
For a permutation $\sigma: \{1,2,\dots, K-1\} \rightarrow \{1,2,\dots, K-1\}$, we denote $\mathcal{S}_\sigma = \{\bold x \in \mathbb{R}^{K-1}: 0\leq x_{\sigma(1)} \leq x_{\sigma(2)} \leq \dots \leq x_{\sigma(K-1)} \leq 1\}$. We can then partition (up to intersections of Lebesgue measure zero) the cube using the $(K-1)!$ different permutations:
\begin{equation}\label{part}
\mathcal{R} = \bigcup\limits_{\sigma} \mathcal{S}_{\sigma},
\end{equation}
where the union is over all permutations. 
While our sample space $\text{cl}(\mathbb{S}^{K-1})$, is not equal to $\mathcal S_\sigma$ for any $\sigma$, we will see in section \ref{sec:eq} that sampling from $\text{cl}(\mathbb{S}^{K-1})$ and sampling from $\mathcal{S}_{id}$ are equivalent, where $id$ is the identity permutation. 
However, as we will see in section \ref{seq:rej_eff}, sampling from $\mathcal{S}_{id}$ allows to take advantage of the cube partitioning of equation \ref{part}, while the same cannot be done for $\text{cl}(\mathbb{S}^{K-1})$.

\subsubsection{The equivalence of sampling over any simplex}\label{sec:eq}

In this section, we consider varying the support of our CC density from the standard simplex, to other simplexes as well as the unit cube.
We denote the support explicitly by writing $\bold x \sim \mathcal{CC}_{\mathcal{A}}(\boldsymbol \eta)$ for the density:
\begin{align}
p_{\mathcal{A}}(\bold x|\boldsymbol \eta) &\propto \exp \left( \sum_{i=1}^{K-1} \eta_i x_i \right) \mathbbm{1}(\bold x \in \mathcal{A})
\end{align}
where the subscript $\mathcal{A}$ will typically denote a simplex.
Now, letting $\bold x \sim \mathcal{CC}_{\mathcal{A}}(\boldsymbol \eta)$ and $\bold y =Q \bold x$, where $Q\in \mathbb{R}^{(K-1)\times (K-1)}$ is an invertible matrix, it follows by the change of variable formula that:
\begin{align} \nonumber
p_{Q(\mathcal{A})}(\bold y |\boldsymbol \eta) & = \dfrac{1}{|\text{det}(Q)|}p_{\mathcal{A}}(Q^{-1}\bold y |\boldsymbol \eta) \nonumber
\\ \nonumber
&\propto \exp(\boldsymbol \eta^\top [Q^{-1} \bold y]) \mathbbm{1}(y \in Q(\mathcal{A})) \\ 
& = \exp([Q^{-\top} \boldsymbol \eta ]^\top \bold y) \mathbbm{1}(y \in Q(\mathcal{A})), 
\end{align}
where $Q(\mathcal{A})=\{\bold y: \bold y=Q\bold x, \bold x\in \mathcal{A}\}$. 
Thus, we have that $\bold y \sim \mathcal{CC}_{Q(\mathcal{A})}(\text{\boldmath$ \tilde \eta$})$, where $ \text{\boldmath$ \tilde \eta$} = Q^{-\top} \boldsymbol \eta$, so that $\bold y$ has a new CC distribution on a transformed sample space.
Moreover, if $Q$ is a permutation matrix and $\mathcal{A} = \mathcal{S}_\sigma$ for some permutation $\sigma$, then $Q(\mathcal{A})$ is a `permuted' simplex, and $\text{\boldmath$ \tilde \eta$}$ is a rearranged parameter vector, hence the equivalence of sampling over any simplex for the CC.


\subsubsection{The permutation sampling algorithm}\label{seq:rej_eff}

Now, consider a lower triangular matrix of ones:
\begin{equation}
B = \begin{pmatrix}
1 & 0 & 0 & \cdots &  0  \\
1 & 1 & 0 & \cdots &  0 \\
1 & 1 & 1 & \cdots &  0 \\
\vdots & \vdots & \vdots & \ddots &  \vdots \\
1 & 1 & 1 & \cdots &  1 \\
\end{pmatrix}
.
\end{equation}
Note that $\mathcal{S}_{id} = B(\text{cl}(\mathbb{S}^{K-1}))$, so that sampling from $\mathcal{CC}_{\text{cl}(\mathbb{S}^{K-1})}(\boldsymbol \eta)$ is equivalent to sampling from $\mathcal{CC}_{\mathcal{S}_{id}}(\text{\boldmath$ \tilde \eta$})$ and transforming the result with $B^{-1}$, where $\text{\boldmath$ \tilde \eta$}= B^{-\top} \boldsymbol \eta$. 
Now consider rejection sampling to draw from $\mathcal{CC}_{\mathcal{S}_{id}}(\text{\boldmath$ \tilde \eta$})$.
As with the naive sampler, our proposal can be drawn on the whole unit cube from independent 1-dimensional CB variates, but the advantage here is that we do not have to directly reject the sample if it fell outside of the desired simplex $\mathcal{S}_{id}$, but rather we can transform it onto that simplex and then accept it with an appropriate probability (which we can compute easily using the invariance property).
Here, the acceptance probability depends on which simplex the proposal fell into, and is given by:
\begin{equation}\label{eq:acc_prob}
\alpha(\bold y, \boldsymbol{ \tilde \eta}, P) = \dfrac{p_{\mathcal{S}_{id}}(\bold y| \boldsymbol{\tilde \eta})}{\kappa(\boldsymbol{\tilde \eta},P)p_{\mathcal{S}_{id}}(\bold y| P^{-\top} \boldsymbol{\tilde \eta})}
,
\end{equation}
where $\kappa(\boldsymbol{\tilde \eta},P)$ is the rejection sampling constant, which in this case is equal to:
\begin{align}\label{eq:opt}
\kappa(\boldsymbol{\tilde \eta},P) = \max_{\bold y \in \mathcal{S}_{id}} \dfrac{p_{\mathcal{S}_{id}}(\bold y| \boldsymbol{\tilde \eta} )}{p_{\mathcal{S}_{id}}(\bold y| P^{-\top} \boldsymbol{\tilde \eta} )}
.
\end{align}

\begin{algorithm}
   \caption{Permutation sampler}
   \label{alg:perm_sampling}
   {\bfseries Input:} target distribution $\mathcal{CC}(\text{\boldmath$ \eta$})$.
   \\
   {\bfseries Output:} sample $\bold x$ drawn from target.
   \\ \vspace{-4mm}
   \begin{algorithmic}[1]
   \STATE Sample $\bold y' \sim \mathcal{CC}_{\mathcal{R}}(\text{\boldmath$ \tilde \eta$})$  (again, this is straightforward to do by sampling each coordinate independently).
   \STATE Sort the elements of $\bold y'$. In other words, find a permutation $\sigma$ such that $\sigma(\bold y') \in \mathcal{S}_{id}$. Let $P$ be the corresponding permutation matrix and $\bold y = P \bold y'$. 
   \STATE Compute $\kappa(\boldsymbol{\tilde \eta},P)$ by taking the maximum over the the vertices of $\mathcal S_{id}$, and use this to compute the acceptance probability $\alpha(\bold y, \boldsymbol{ \tilde \eta}, P)$.
   \STATE Accept $\bold y$ with probability $\alpha(\bold y, \boldsymbol{ \tilde \eta}, P)$. Otherwise, go back to step 1.
   \STATE Return $\bold x = B^{-1} \bold y$.
\end{algorithmic}
\end{algorithm}

The algorithm samples correctly from $\mathcal{CC}_{\mathcal{S}_{id}}(\text{\boldmath$ \tilde \eta$})$, because $\bold y'$ can be thought of as a sample from $p_{\mathcal{R}}(\bold y'|\boldsymbol \eta, \bold y'\in \mathcal{S}_{\sigma})=p_{\mathcal{S}_{\sigma}}(\bold y'|\boldsymbol \eta)$, which we then transform with $P$ to obtain a distribution on $\mathcal{S}_{id}$. If we use this distribution as a proposal distribution for a rejection sampling algorithm, we recover precisely the acceptance probability of equation \ref{eq:acc_prob}. Intuitively, if our sample $\bold x$ does not fall on the desired simplex $\mathcal{S}_{id}$, we move around the simplex in which it fell (along with $\bold x$ itself) so that it matches the desired simplex, and then do rejection sampling. 

We conclude this subsection with two short notes on the optimization problem of equation \ref{eq:opt}. 
The first one is that when $\boldsymbol \eta^{(2)} = Q^{-\top} \boldsymbol \eta^{(1)}$ where $|\text{det}(Q)|=1$, then the normalizing constants cancel out, which is the case in our algorithm since $|\text{det}(P)|=1$. 
The second is that, by taking logs, the optimization problem can be transformed into a linear problem subject to linear inequality constraints, meaning that the solution must be achieved at a vertex. 
Since there are $K$ vertices, namely $\bold{0}$, $\bold{e}_{K-1}, \bold{e}_{K-1} + \bold{e}_{K-2}, \dots, \sum_{i=1}^{K-1} \bold{e}_i$, we can solve the problem by simply checking each of these vertices.

\subsection{Performance}

\begin{figure}
\centering
  \includegraphics[width=0.7\linewidth]{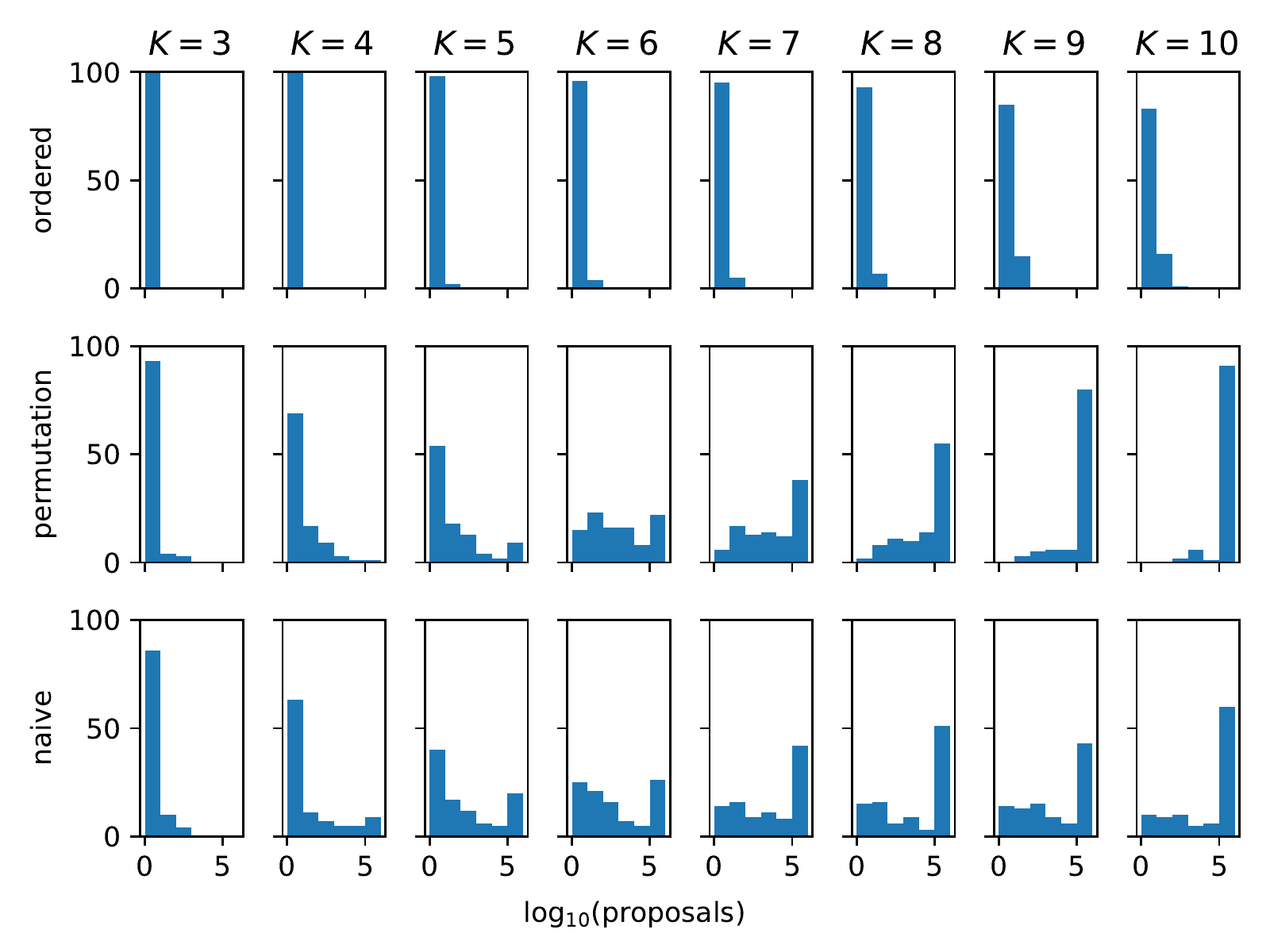}
  \caption{Shows the performance of 3 sampling algorithms across different dimensions $K$. Each histogram shows the distribution, over 100 trials, of the number of proposals required for 1 acceptance, on the log scale (base 10). The distributions are not exponential, since each of the 100 trials is sampled from a different $\mathcal{CC}(\text{\boldmath$  \lambda$})$ distribution, where the parameter follows $\text{\boldmath$  \lambda$} \overset{iid}{\sim} Dirichlet(1/K,\cdots,1/K)$. Due to computational constraints, the number of proposals in each trial is right-censored, hence the large bars at the right end of the histograms.}
  \label{fig:rejection_rates_hist}
\end{figure}

While the ordered rejection sampler can never have a worse rejection rate than its naive counterpart, the comparison with the permutation sampler depends on the shape of the target distribution, as discussed.
The perfectly balanced case $\text{\boldmath$  \lambda$}=(1/K,\dots,1/K)$ results in the worst possible rejection rate for the ordered rejection sampler (we accept with probability $1/(K-1)!$), but also the best possible rejection rate for the permutation sampler (this is the uniform case so $\alpha(\bold y,\boldsymbol{\tilde \eta}, P) = 1$).
On the other end of the spectrum, in the totally unbalanced case where one element of $\text{\boldmath$  \lambda$}$ holds all the weight and the others are close to zero, the ordered rejection sampler achieves an acceptance rate close to 1, whereas it is much smaller for the permutation sampler (see section \ref{sec:rej_rates_ex}).
In this sense, our samplers are complementary, and an optimal sampling algorithm could involve combining accept/reject steps from both methods.
We study the performance of our samplers empirically, by comparing the distribution of the rejection rates under a sparsity-inducing prior $\text{\boldmath$  \lambda$} \sim Dirichlet(1/K,\dots,1/K)$.
Indeed, the ordered rejection sampler tends to considerably outperform the permutation sampler (see figure \ref{fig:rejection_rates_hist}), as well as (trivially) the naive sampler.

Now, it may come as a surprise that the permutation sampler does not necessarily outperform the naive sampler.
After all, the naive sampler only accepts samples that fell directly into the desired simplex, whereas the permutation sampler has the additional possibility of accepting a sample that fell outside of $\mathcal{S}_{id}$ after applying a suitable permutation.
However, this intuition breaks down once we realize that the proposal distributions from the two methods are not equivalent. We make this precise in the following sections.




\subsubsection{Rejection rate - naive sampler}

Suppose we seek $\bold x\sim \mathcal{CC}_{\text{cl}(\mathbb{S}^{K-1})} (\boldsymbol \eta)$. The naive rejection sampler proposes $\bold x \sim \mathcal{CC}_{\mathcal{R}} (\boldsymbol \eta)$, and accepts if $\bold x \in {\mathbb{S}^{K-1}}$. The proposal density is equal to (we know the normalizing constant as we have the product of independent CBs):
\begin{align}
p_{\mathcal{R}}(\bold x|\boldsymbol \eta) = \prod_{i=1}^{K-1} \frac{{\eta_i}}{e^{\eta_i} -1} e^{\eta_i x_i}.
\end{align}
Therefore, the probability of acceptance is:
\begin{align}
P(\mathcal{CC}_{\mathcal{R}}(\boldsymbol \eta) \in {\mathbb{S}^{K-1}}) = \int_{\mathbb{S}^{K-1}} \prod_{i=1}^{K-1} \frac{{\eta_i}}{e^{\eta_i} -1} e^{\eta_i x_i} d\mu.
\end{align}
We can apply the transformation $B$ to rewrite this as:
\begin{align}
P(B(\mathcal{CC}_{\mathcal{R}}(\boldsymbol \eta)) \in B({\mathbb{S}^{K-1}}) ) = P(\mathcal{CC}_{B(\mathcal{R})}(B^{-\top} \boldsymbol \eta) \in S_{id}) = \int_{\mathcal{S}_{id}} \prod_{i=1}^{K-1} \frac{{\eta_i}}{e^{\eta_i} -1} e^{\tilde \eta_i x_i}  d\mu,
\end{align}
where we have used the fact that $|\det(B)| = 1$ so that the normalizing constant remains unchanged. Thus, the probability of acceptance of the naive sampler is equal to:
\begin{align} \label{naive_acceptance_rate}
P(accept) =  \left( \prod_{i=1}^{K-1} \frac{{\eta_i}}{e^{\eta_i} -1} \right) \cdot \int_{\mathcal{S}_{id}}  e^{\text{\boldmath$ \tilde \eta$}^\top \bold x} d\mu.
\end{align}

\subsubsection{Rejection rate - permutation sampler}

In the case of the permutation sampler, the acceptance rate is harder to compute. However, one can easily obtain a lower bound by considering only the samples that fall directly into our target simplex (in this case, $\mathcal{S}_{id}$).
In this case, the proposal distribution is:
\begin{align}
p_{\mathcal{R}}(\bold x| \text{\boldmath$ \tilde \eta$}) = \prod_{i=1}^{K-1}  \frac{{\tilde\eta_i}}{e^{\tilde\eta_i} -1} e^{\tilde\eta_i x_i}.
\end{align}
If the resulting sample falls in $\mathcal S_{id}$, we accept the sample and map it back to $\text{cl}({\mathbb{S}^{K-1}})$. Thus, the acceptance rate has the lower bound
\begin{align}
P(accept) \ge \left( \prod_{i=1}^{K-1}  \frac{{\tilde\eta_i}}{e^{\tilde\eta_i} -1} \right) \cdot \int_{\mathcal{S}_{id}}  e^{\text{\boldmath$ \tilde \eta$}^\top \bold x} d\mu.
\end{align}
Note that, while this lower bound has the same integral term as the naive rejection sampler, it is multiplied by a different normalizing constant. In particular, there are configurations of $\boldsymbol \eta$ that can lead to much worse normalizing constants for the permutation sampler than for the naive rejection sampler, resulting in a worse acceptance rate overall.

\subsubsection{Example} \label{sec:rej_rates_ex}

We give an example of a configuration of $\boldsymbol \eta$ such that the acceptance rate of the naive sampler is better than that of the permutation sampler.
Consider the case $(\eta_1, \cdots, \eta_{K-1}) = (-M, \cdots, -M)$, where $M$ is a large positive number. 
Note that this example is far from the uniform case $\text{\boldmath$  \lambda$} = (1/K, \cdots, 1/K)$.
In fact, in this case, after transforming with $B$, we obtain $\tilde \eta_{K-1} = -M$, and $\tilde \eta_{K-2} = \cdots = \tilde \eta_1 = 0$.
Thus, when we sample a proposal $\bold y \sim \mathcal{CC}_{\mathcal R} ( \text{\boldmath$ \tilde \eta$})$, typically $y_{K-1}$ will be small relative to $y_1, \cdots, y_{K-2}$, which will be ordered at random.
This means the sorting step of our permutation sampler will likely map $y_{K-1}$ to $y'_1$, as well as sorting the remaining entries $y'_2<\cdots < y'_{K-1}$.
In other words, $P$ maps the $(K-1)^{th}$ entry to the 1st entry, and one of the first $K-2$ entries to the $(K-1)^{th}$ entry (whichever of these happens to sample the largest value).
The resulting distributions $p_{\mathcal S_{id}}(\cdot |  \text{\boldmath$ \tilde \eta$})$ and $p_{\mathcal S_{id}} ( \cdot | P^{-\top}  \text{\boldmath$ \tilde \eta$})$ are similar, with the key difference that the former puts the negative $ \text{\boldmath$ \tilde \eta$}$ coefficient (namely $\tilde \eta _{K-1} = -M$) in the last position (the largest component of $\bold y$), while the latter puts it into some other position determined by $\sigma^{-1}$, i.e. the right-most column of $P$ or equivalently, the bottom row of $P^{-1}$.
In this setting, it follows that $\kappa$ will be equal to 1, and the ratio $p_{\mathcal S_{id}}(\bold y' | \text{\boldmath$ \tilde \eta$}) / p_{\mathcal S_{id}} ( \bold y' | P^{-\top} \text{\boldmath$ \tilde \eta$})$ will be small.
Thus, our rejection sampling ratio is typically small and we are likely to reject our proposal.
The ratio will be close to 1 only in the event that the proposed value $x_{K-1}$ is large relative to the other components $x_1,\cdots,x_{K-2}$, which rarely happens as $x_{K-1}$ is sampled from a univariate CC with much smaller coefficient.
Note further, that $\text{\boldmath$ \tilde \eta$}$ cannot be re-shuffled in this case, as this would lead to a target distribution on a simplex other than $\mathcal S_{id}$ (we can only shuffle $\boldsymbol \eta$ prior to applying $B$, which in this case leaves $\text{\boldmath$ \eta$}$ unchanged). 
We conclude that we cannot achieve a uniformly better rejection rate through the permutation sampler, relative to the naive method.

%
%



\bibliography{ref}
\bibliographystyle{icml2020}